\documentclass[10pt,journal,compsoc]{IEEEtran}

\ifCLASSOPTIONcompsoc
  \usepackage[nocompress]{cite}
  \usepackage[caption=false,font=footnotesize,labelfont=sf,textfont=sf]{subfig}
\else
  \usepackage{cite}
  \usepackage[caption=false,font=footnotesize]{subfig}
\fi
\ifCLASSINFOpdf
  \usepackage[pdftex]{graphicx}
  \DeclareGraphicsExtensions{.pdf,.jpeg,.png}
\else
  \usepackage[dvips]{graphicx}
  \DeclareGraphicsExtensions{.eps}
\fi
\usepackage{amsmath,amssymb,amsfonts,dsfont,pifont,bm,bbm,mathrsfs,mathtools}
\usepackage{algorithm,algpseudocode,listings}
\usepackage{booktabs,multirow,adjustbox,diagbox}
\usepackage{xspace}
\usepackage{float,capt-of}
\usepackage[table]{xcolor}
\usepackage{url}
\usepackage[pagebackref=true,breaklinks=true,colorlinks=true,citecolor=blue,bookmarks=false]{hyperref}
\usepackage[capitalize]{cleveref}  
\crefname{section}{Sec.}{Secs.}
\Crefname{section}{Section}{Sections}
\crefname{table}{Tab.}{Tabs.}
\Crefname{table}{Table}{Tables}
\crefname{figure}{Fig.}{Figs.}
\Crefname{figure}{Figure}{Figures}
\crefname{equation}{Eq.}{Eqs.}
\Crefname{equation}{Equation}{Equations}
\newcommand{\Loss}{\mathcal{L}}     

\newcommand{\etal}{\textit{et al. }}

\newcommand{\tocite}[1]{\textcolor{red}{[TOCITE]}}

\begin{document}

\newcommand{\titlename}{Image Synthesis under Limited Data: A Survey and Taxonomy}
\title{Image Synthesis under Limited Data: A Survey and Taxonomy}

\author{
Mengping Yang,
Zhe Wang*
\IEEEcompsocitemizethanks{
    \IEEEcompsocthanksitem M. Yang and Z. Wang are with  Key Laboratory of Smart Manufacturing in Energy Chemical Process, Ministry of Education, East China University of Science and Technology, Shanghai, 200237, P. R. China.  Department of Computer Science and Engineering, East China University of Science and Technology, Shanghai, 200237, P. R. China. \protect \\
    E-mail: wangzhe@ecust.edu.cn, mengpingyang@mail.ecust.edu.cn
  }%
  \thanks{* corresponding author.}
}


\IEEEtitleabstractindextext{
    \begin{abstract}
Deep generative models, which target reproducing the given data distribution to produce novel samples, have made unprecedented advancements in recent years.
Their technical breakthroughs have enabled unparalleled quality in the synthesis of visual content.
However, one critical prerequisite for their tremendous success is the availability of a sufficient number of training samples, which requires massive computation resources.
When trained on limited data, generative models tend to suffer from severe performance deterioration due to overfitting and memorization.
Accordingly, researchers have devoted considerable attention to develop novel models that are capable of generating plausible and diverse images from limited training data recently.
Despite numerous efforts to enhance training stability and synthesis quality in the limited data scenarios, there is a lack of a systematic survey that provides
1) a clear problem definition, critical challenges, and taxonomy of various tasks;
2) an in-depth analysis on the pros, cons, and remain limitations of existing literature;
as well as 3) a thorough discussion on the potential applications and future directions in the field of image synthesis under limited data.
In order to fill this gap and provide a informative introduction to researchers who are new to this topic, this survey offers a comprehensive review and a novel taxonomy on the development of image synthesis under limited data.
In particular, it covers the problem definition, requirements, main solutions, popular benchmarks, and remain challenges in a comprehensive and all-around manner.
We hope this survey can provide an informative overview and a valuable resource for researchers and practitioners, and promote further progress and innovation in this important topic.
Apart from the relevant references, we aim to constantly maintain a timely up-to-date repository to track the latest advances in this topic at \href{https://github.com/kobeshegu/awesome-few-shot-generation}{GitHub/awesome-few-shot-generation}.

\end{abstract}


    \begin{IEEEkeywords}
        Generative modeling, limited data, few-shot image generation, data-efficiency, generative domain adaptation.
    \end{IEEEkeywords}
}

\maketitle
\IEEEdisplaynontitleabstractindextext
\IEEEpeerreviewmaketitle

\IEEEraisesectionheading{\section{Introduction}
\label{sec:introduction}}

\IEEEPARstart{D}{eep} generative models have made tremendous development and have been applied to a wide range of intelligent creation tasks, particularly in image and video composition~\cite{bigan, stylegan2, stylegan3, styleganv, latentdiffusion, zhang2022towards, dhariwal2021diffusion, yan2021videogpt, ho2022imagen}, audio and speech synthesis~\cite{prenger2019waveglow, kong2020diffwave, pascual2023full, huang2022fastdiff, lam2022bddm, kang2023grad}, multi-modal generation~\cite{dalle, dalle2, imagen}, \emph{etc.}
Their technical breakthroughs have also directly facilitate our daily life in many aspects including content creation of various representations (\emph{e.g.,} 3D/2D representations)~\cite{3DGIF, eg3d, graf, 3dshapenet}, customized generation and editing~\cite{kwon2023one, hertz2022prompt, mokady2023null, kumari2023multi, ruiz2023dreambooth}, and artistic synthesis/manipulation~\cite{starganv2, shen2020interfacegan, yang2022pastiche, xu2022transeditor}.
Despite these remarkable advances, most existing generative models require massive amounts of data and computational resources for training. 
For instance, the most commonly used datasets, the human face FFHQ~\cite{FFHQgithub, stylegan2} ($70 K$), the outdoor/indoor scene LSUN~\cite{lsun} ($1 M$), and the object ImageNet~\cite{ImageNet} ($1 M$), all contains sufficient training samples.
Such prerequisite poses a significant challenge for practitioners and researchers who only have limited training samples, like paintings from famous artists and medical images of scarce diseases.
Accordingly, there is an increasing need to learn a generative model under limited training data, which has drawn extensive attention in recent years.

The main challenge of image synthesis under limited data is the risk of model overfitting and memorization, which can significantly affect the fidelity and diversity of the generated samples~\cite{stylegan2-ada, DiffAug, InsGen, projectedGAN, tran2021data}.
Namely, the model might simply duplicate training images instead of generating novel ones due to overfitting, leading to degraded synthesis quality.
For instance, when generative adversarial networks (GANs)~\cite{gan} are trained under limited data, the discriminator is prone to memorize the training images and thus provides meaningless guidance to the generator, resulting in unfavorable synthesis.
In order to address these limitations, many research works have been developed to ameliorate the synthesis quality in the few-shot scenarios~\cite{stylegan2-ada, DiffAug, InsGen, wavegan, wang2022fregan, he2024litegfm}.
These works propose various strategies to mitigate the risk of overfitting and memorization from different perspectives, such as data augmentation, regularization, and novel architectures.

Despite conspicuous progress has been made in the field of image synthesis under limited data, there is a lack of a unified problem definition and taxonomy for this field.
Few-shot image generation, for instance, is defined as producing diverse and realistic images for a unseen category given a few images from this category in~\cite{hong2020f2gan, wavegan, ding2022attribute, ding2023stable}, whereas in~\cite{yang2021one, wang2020minegan, freezeD, ojha2021few, zhao2022few}, few-shot image generation refers to adapting the prior knowledge of a large-scale and diverse source domain to a small target domain.
However, they are significantly different in problem requirements, model training, and testing setups.
This inconsistent definition might lead to ambiguity and misunderstandings among readers who are not familiar with these works.
Therefore, a comprehensive problem definition and taxonomy are vital to facilitate a clearer understanding of this field.
Moreover, considering the lack of a systematic survey and the increasing interest in limited data generation, we believe that it is necessary to organize one to help the community track its development.
To this end, this paper first presents a clear problem definition for various tasks in the few-shot regimes and categorizes them into four categories:
data-efficient generative models (\cref{sec:data_efficient_generative_models}),
few-shot generative adaptation (\cref{sec:few_shot_generative_adaptation}),
few-shot image generation (\cref{sec:few_shot_image_generation}),
and one-shot image synthesis (\cref{sec:one_shot_geneartion}).
%
%
Then, this paper presents an all-around overview of prior studies in this field.
In particular, the technical evolution, advantages, and disadvantages of existing alternatives are presented.
Additionally, we present several related applications and highlight open problems that require further investigation for future works (\cref{sec:applications_future_direction}).

Overall, this survey aims to provide a comprehensive and systematic understanding of image synthesis under limited data for scholars who are new to the field.
Hopefully, our work could serve as a guideline for researchers who are willing to develop their own generative models with only dozens of training images.
The contributions of this survey are summarized in the following:
\begin{itemize}
    \item \textbf{A clear problem definition and taxonomy.} This survey presents a clear and unified problem definition for various synthesis tasks in image synthesis under limited data. Moreover, this survey proposes a systematic taxonomy that divides these tasks into four categories: data-efficient image generation, few-shot generative adaptation, few-shot image generation, and one-shot image synthesis.
    \item \textbf{Comprehensiveness.} This survey provides a comprehensive overview of existing state-of-the-art generative models in the few-shot regimes. 
    We compare and analyze the main technical motivations, contributions, and limitations of existing approaches, which can inspire potential solutions for further improvement.
    \item \textbf{Applications and open research directions.} In addition to the technical investigation, this survey also discusses potential applications and highlights open research problems that require further investigation for the improvement of image synthesis under limited data.
    \item \textbf{A timely up-to-date repository.} In order to continuously track the rapid development of this field, we provide a curated list of the latest relevant papers, code, and datasets at \href{https://github.com/kobeshegu/awesome-few-shot-generation}{GitHub/awesome-few-shot-generation}.
\end{itemize}

The remainder of this paper is organized as follows.
~\cref{sec:scope} presents the scope of this survey and discusses the differences with other surveys.
~\cref{sec:fundamentals} introduces the fundamentals of image synthesis under limited data, namely deep generative models, few-shot learning, and transfer learning.
~\cref{sec:data_efficient_generative_models}, ~\cref{sec:few_shot_generative_adaptation}, ~\cref{sec:few_shot_image_generation}, and~\cref{sec:one_shot_geneartion} respectively provides the detailed comparison and discussions on the existing approaches for the aforementioned tasks.
~\cref{sec:applications_future_direction} discusses the downstream applications and highlights several future research directions.
Finally, ~\cref{sec:conclusion} concludes this survey.


\section{Scope and Overview}
\label{sec:scope}

\noindent \textbf{Scope.}
This survey focuses on methods that train deep generative models to produce diverse and plausible images under limited training data.
The main objective of these approaches is to mitigate the overfitting problem by fully leveraging the internal information of limited training data and producing novel samples within the data distribution.
However, these methods differ in the model input, training diagrams, and evaluation.
Thus, in this survey, we aim to  
1) give readers a clear understanding of various problem settings in the field of image synthesis under limited data, 
2) provide in-depth analysis and insightful discussion about the model concepts, method characteristics, and applications of prior arts,
and 3) pose some research directions for future investigation, and inspire more interesting works for further improvement.
In particular, based on the problem definition and experimental settings, we categorize existing approaches into four groups:
data-efficient generative models,
few-shot generative adaptation,
few-shot image generation,
one-shot image generation.
It is important to note that all these categories aim to synthesize photorealistic and diverse images corresponding to the data distribution. 
This is in contrast to generative modeling in few-shot learning, which explicitly estimates the probability distribution to compute the class label of given samples~\cite{chi2021learning, 9634045}.
Regarding the progress of few-shot learning, we refer readers to~\cite{wang2020generalizing, song2023comprehensive} for a more comprehensive review.

\noindent \textbf{The differences between our survey and others.}
Although there are already some other surveys that discuss the developments, main challenges, potential applications, and future opportunities of deep generative models~\cite{shi2022deep, gui2021review, jabbar2021survey, 10081412, liu2021generative}, 
very few have focused on the development of deep generative models in limited data scenarios.
The most relevant work to ours is~\cite{li2022comprehensive}, which analyzes the degradation of data-efficient GANs and provides a novel taxonomy and opportunities.
However, our survey has several advantages over~\cite{li2022comprehensive}:
1) \textit{Comprehensive investigation.}
In~\cite{li2022comprehensive}, only the traditional noise-to-image scheme of data-efficient GANs is discussed. 
In contrast, conditional generative models that use few conditional images~\cite{hong2020f2gan, matchinggan, wavegan} as input are also taken into account in this survey.
Besides, this survey covers one-shot image synthesis tasks that are solely trained on single image~\cite{singan, oneshotgan, zhang2021exsingan} and few-shot image generation tasks that produce novel samples for a category given few images from the same category, whereas~\cite{li2022comprehensive} ignores.
2) \textit{Up-to-date investigation.}
Since the publication of \cite{li2022comprehensive}, the limited-data synthesis field has seen significant progress, such as the integration of diffusion models~\cite{wang2022sindiffusion, giannone2022few, kulikov2023sinddm} and the consideration of inversion techniques~\cite{inversionsurvey, zhu2020domain, wang2022high, bai2022high}. 
Our survey provides a timely and up-to-date review of these recent advances.
3) \textit{Thorough discussion and analysis.}
Our survey provides a more detailed comparison of the design concepts, model details, and method characteristics of existing approaches. 
Additionally, we highlight the potential applications in practical domains and the technical limitations that require further investigation.
In brief, this survey covers all related works presented in~\cite{li2022comprehensive} while containing the most recent advances and more comprehensive investigations.
Notably, our survey complements existing overviews of generative models by providing a comprehensive and systematic understanding of image synthesis under limited data.

\begin{figure*}[t]
    \centering
    \includegraphics[width=0.95\textwidth]{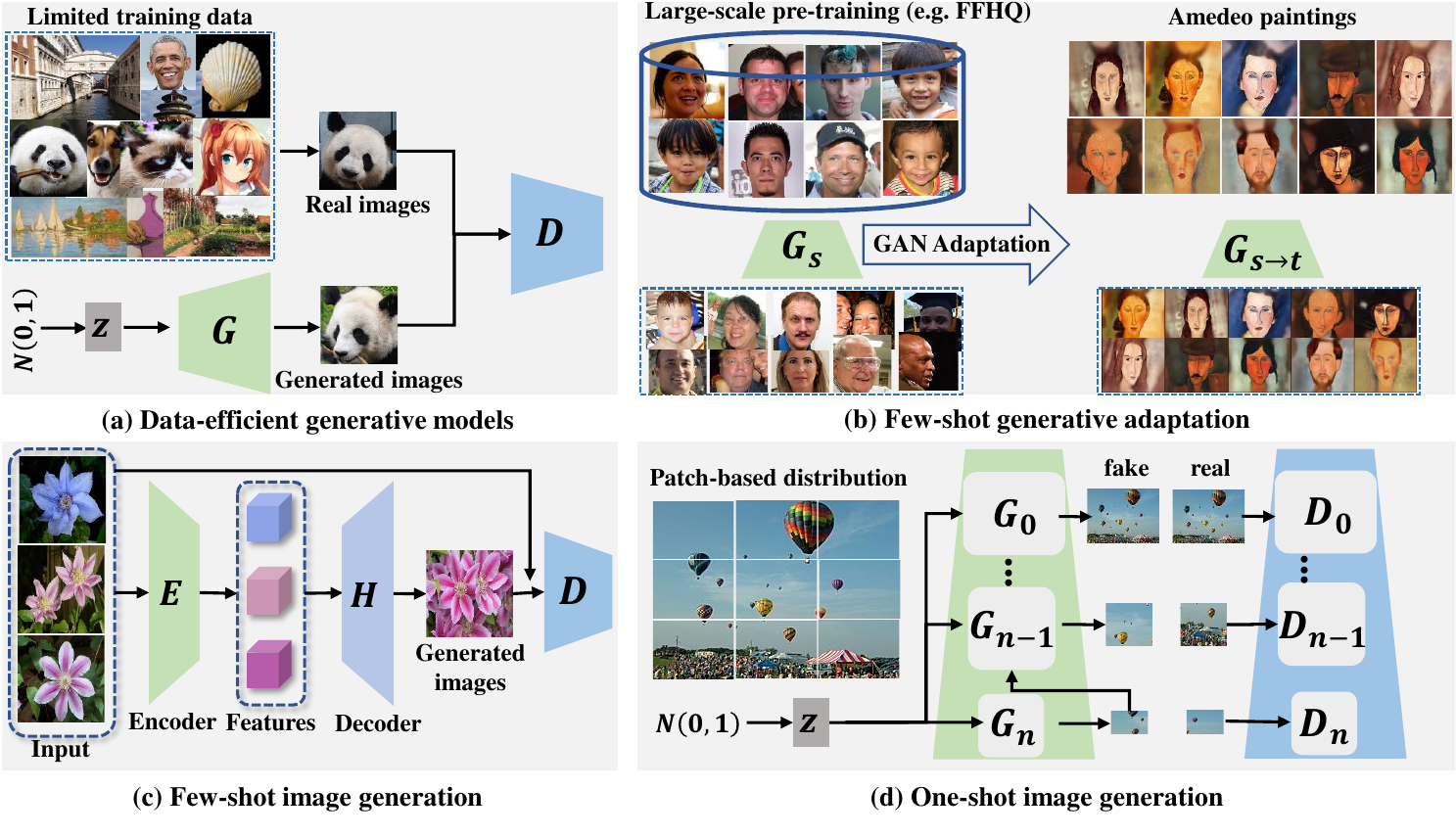}
    \caption{Various problem settings of image synthesis models under limited data. 
    In particular,
    (a) represents the data-efficient GAN training pipeline that learns to capture the observed distribution from scratch with limited data;
    (b) denotes the pipeline of few-shot generative adaptation, which transfers prior knowledge from pre-trained large-scale source generative models to target domains with very few images, \emph{e.g.,} 10-shot images;
    (c) shows the learning scheme of few-shot image generation, the model is expected to produce novel samples given several input conditional images;
    (d) presents the training process of one-shot image generation, the model is trained solely on one single image in a coarse-to-fine manner to capture the underlying internal distribution of the reference image.
    It is imperative to note that the sub-figures presented herein serve solely as illustrative aids to convey the problem settings of diverse image synthesis tasks.
    As such, it should be understood that not all approaches utilized in these tasks are consistent with the pipeline depicted in the diagrams.
    }
    \label{fig:problem_definition}
\end{figure*}

\noindent \textbf{Overview.}
In this survey, we aim to provide a lucid comprehension of various tasks concerning image synthesis under limited data.
To achieve this goal, we present the definition and formulation of each task, taking into account the training paradigms and task-specific requirements that underlie each problem. 
The four independent problems that we have formulated are data-efficient generative models, few-shot generative adaptation, few-shot image generation, and one-shot image generation.
In order to better illustrate these problems, we consider one representative category in the family of deep generative models, namely Generative Adversarial Networks (GANs), to depict the training pipelines of these problems in~\cref{fig:problem_definition}. 
It is important to note that the presented pipeline is not intended to represent all approaches utilized in each task, but rather serves as an exemplar.
Moreover, we summarize the definitions, model requirements, and primary challenges of each task in~\cref{tab:taxonomy_challeges}.
The detailed methodology design and taxonomy are presented respectively in the corresponding sections.

\begin{table*}
	\center
	\caption{Problem categories, formulation, key challenges, and primary solutions of various tasks for image synthesis under limited data.}
	\label{tab:taxonomy_challeges}
	\resizebox{\linewidth}{!}{
	\small
		\begin{tabular}{llll}
			\toprule
			Problem Categories & Problem Formulation & Challenges & Key Solutions  \\
			\midrule
			Data-efficient Generative Models & Directly trained on $\mathbf{D}$
            & \begin{tabular}[l]{@{}l@{}}Model overfitting\\Model memorization\\Unstable training\end{tabular} & \begin{tabular}[l]{@{}l@{}}Data Augmentation\\Architecture-variants\\Loss Regularization\end{tabular} \\
			\hline
			Few-shot Generative Adaptation & Transfer from $\mathbf{D_s}$ to $\mathbf{D_t}$
            & \begin{tabular}[l]{@{}l@{}}Model overfitting\\Domain gaps\\Incompatible knowledge\end{tabular} & \begin{tabular}[l]{@{}l@{}}Fine-tuning\\Introducing extra branches\\Loss regularization\end{tabular} \\
            \hline
			Few-shot Image Generation & Learn to generate $\mathbf{D_u}$ from $\mathbf{D_s}$
            & \begin{tabular}[l]{@{}l@{}}Catastrophic forgetting\\Knowledge transfer\\Fail to generalize\end{tabular} & \begin{tabular}[l]{@{}l@{}}Optimization-based\\Transformation-based\\Fusion-based\\Inversion-based\end{tabular} \\
            \hline
			One-shot Image Generation & Learn the internal distribution of a single image 
            & \begin{tabular}[l]{@{}l@{}}Collapse to replicate input image\\Synthesis variance\\Training time\end{tabular} & \begin{tabular}[l]{@{}l@{}}Multi-stage training\\Patch-based training\\Distribution matching\end{tabular} \\
			\bottomrule
		\end{tabular}
	}
\end{table*}

\section{Fundamentals}
\label{sec:fundamentals}
This section briefly introduces the fundamentals of image synthesis under limited data, encompassing deep generative models, few-shot learning, and transfer learning. 
The general concepts and overall pipelines of these fundamentals are presented, enabling readers to gain a preliminary understanding of our survey. 
Notably, the detailed taxonomy and model designs of these fundamentals are beyond the scope of this section.
For a more comprehensive survey of few-shot learning and transfer learning, we refer readers to~\cite{wang2020generalizing, song2023comprehensive} and~\cite{zhuang2020comprehensive, niu2020decade}.

\begin{figure}[t]
    \centering
    \includegraphics[width=0.48\textwidth]{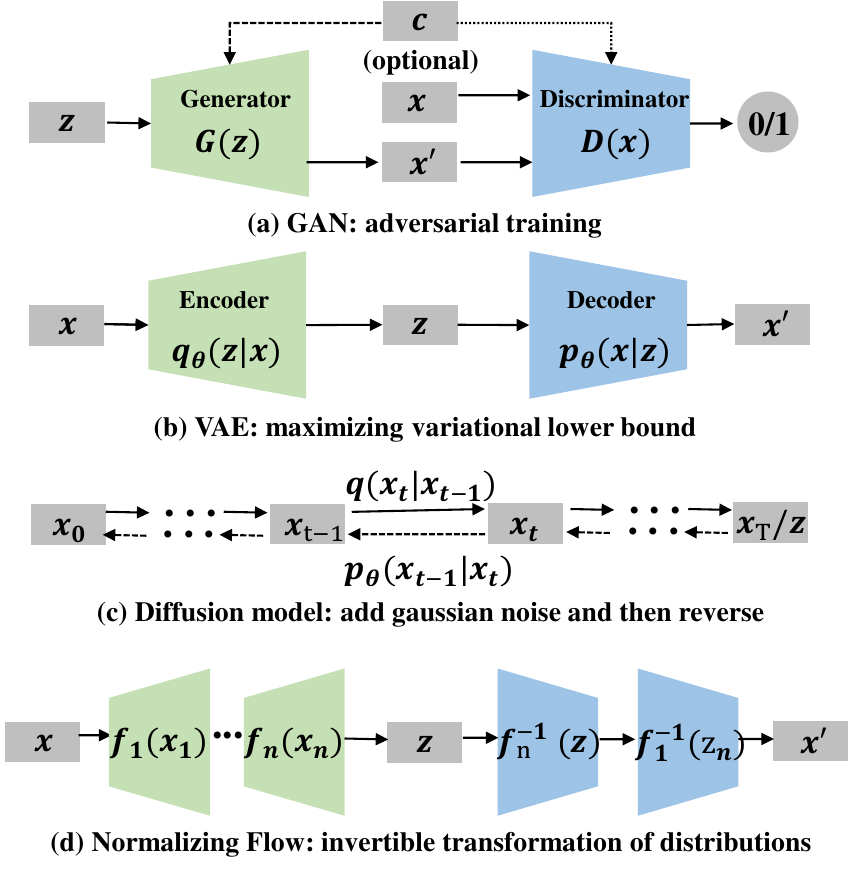}
    \caption{An overview of various deep generative models.}
    \label{fig:generative_models}
\end{figure}

\subsection{Deep Generative Models}
Generative models aim to capture the actual underlying distribution of a given set of training data, with the goal of generating novel samples that closely follow this distribution. 
To achieve this, generative models must capture fine-grained details as much as possible.
Typically, existing generative models can be broadly categorized into two categories: likelihood-based models and likelihood-free models.
Likelihood-based models explicitly maximize the likelihood probability of the given data distribution, 
while likelihood-free models implicitly capture the data distribution through a mini-max game between two sub-networks.
In particular, likelihood-based generative models could be further categorized into Variational AutoEncoders (VAEs)~\cite{vae}, Diffusion Probabilistic Models (DPMs)~\cite{sohl2015deep, dhariwal2021diffusion}, and Normalizing Flows (NFs)~\cite{normalizingflows}, whereas the likelihood-free one usually refers to Generative Adversarial Networks (GANs)~\cite{gan}.
~\cref{fig:generative_models} presents the overall pipeline of each generative model.
In the following, we briefly introduce the formulations and concepts underlying each generative model.

\noindent \textbf{Generative Adversarial Networks (GANs).}
Generative Adversarial Networks (GANs) typically consist of two sub-networks, a generator and a discriminator.
The two sub-networks are trained in an adversarial manner, where the generator tries to fool the discriminator by producing images that are difficult to distinguish from real ones, 
and the discriminator tries to correctly identify whether an image is real or generated.
This process is optimized with a two-player mini-max game between the generator and the discriminator in an adversarial manner.
Formally, they are optimized by
\begin{equation}
\begin{aligned}
\mathcal{L}_D & =-\mathbb{E}_{\mathbf{x} \sim p_{\mathbf{x}}}[\log (D(\mathbf{x, c}))]-\mathbb{E}_{\mathbf{z} \sim p_{\mathbf{z}}}[\log (1-D(G(\mathbf{z}, c)))], \\
\mathcal{L}_G & =-\mathbb{E}_{\mathbf{z} \sim p_{\mathbf{z}}}[\log (D(G(\mathbf{z, c})))].
\end{aligned}
\label{eq:GAN_equation}
\end{equation}
where $D(\cdot)$ and $G(\cdot)$ denote the discriminator and the generator, respectively.
$\mathbf{x}$ represents the real samples and $G(\mathbf{z})$ denotes the generated ones.
$\mathbf{c}$ is the additional condition for conditional geneartion, which is optional during training.
Targeting at reaching a Nash equilibrium between the generator and the discriminator, GANs are notoriously difficult to train.
As a result, several issues like gradient vanishing, mode collapse, and training divergence are prone to happen. 
In order to improve the training stability and model performances of GANs, enormous research efforts have been poured, mostly focus on designing loss-variant and architecture-variant models.
For instance, PG-GAN~\cite{pggan}, BigGAN~\cite{bigan}, and the StyleGAN series~\cite{stylegan, stylegan2, stylegan3} have developed dedicated architectures that enable the injection of fine-details into the generating process, resulting in photo-realistic image synthesis.
In contrast, WGAN~\cite{arjovsky2017wasserstein, gulrajani2017improved} and f-GAN~\cite{nowozin2016f} employ different optimization objectives to ameliorate the generation quality.
Conditional GANs have been formulated to enable more controllable generation by injecting additional conditions (such as class labels~\cite{mirza2014conditional, odena2017conditional, ho2021classifier}, segmentation masks~\cite{park2019semantic, schonfeld2021you, luo2022context}, or training images~\cite{casanova2021instance, wang2018high}) into both the generator and the discriminator.
However, many of the previous GANs are trained on large-scale datasets.
When the training data is limited, problems like overfitting~\cite{stylegan2-ada, jiang2021deceive}, memorization~\cite{DiffAug, InsGen} and non-convergent training~\cite{liu2020towards} might arise.

\noindent \textbf{Variational Autoencoders (VAEs).}
Variational AutoEncoders (VAEs) aim to learn a latent variable model that captures the underlying distribution of the data, thereby enabling the learning of a compressed representation of the data that captures its essential features.
In other words, VAEs seek to learn a probabilistic mapping from the input data to the latent space.
VAEs are trained to maximize the likelihood of the data while simultaneously minimizing the distance (\emph{i.e.,} KL divergence) between the learned latent variable distribution and a prior distribution, \emph{e.g.,} a standard normal distribution.
This is achieved by optimizing a variational lower bound on the log-likelihood of the data, which consists of two terms: the reconstruction loss and the KL divergence loss.
Formally, VAEs minimize a variational lower bound on the log-likelihood of the training data:
\begin{equation}
\mathcal{L}_{V A E}=-D_{K L}\left(q_\phi(\mathbf{z} \mid \mathbf{x}) \| p_\theta(\mathbf{z})\right)+\mathbb{E}_{\mathbf{z} \sim q_{\phi(\mathbf{z} \mid \mathbf{x})}} \log \left(p_\theta(\mathbf{z} \mid \mathbf{x})\right).
\label{eq:vaes}
\end{equation}
where $q_\phi(\mathbf{z} \mid \mathbf{x})$ denotes the approximation of posterior probability,
and $p_\theta(\mathbf{z} \mid \mathbf{x})$ represents log-likelihood of the training data.
Variational AutoEncoders (VAEs) are trained to minimize Eq.~\ref{eq:vaes} with respect to the parameters of the encoder and decoder neural networks.
Once trained, the decoder of VAEs can be used to generate new samples by sampling from the prior distribution and decoding the resulting latent variables.
Compared to GANs, VAEs are more stable to optimize due to the use of log likelihood estimation.
However, the synthesized samples of VAEs are often blurry and noisy due to the injection of noise distribution and imperfect pixel-level reconstruction.
Moreover, the imbalance between the prior distribution (\emph{i.e.,} a Gaussian) and the limited training data may make VAEs difficult to optimize, leading to unsatisfactory synthesis performance and unstable training, particularly in few-shot scenarios.
Therefore, VAEs are not suitable for image synthesis under limited data.
%
%

\noindent \textbf{Diffusion Probabilistic Models (DPMs).}
The basic idea of DPMs is to learn a stochastic process that can transform a pure distribution of noise, \emph{i.e.,} Gaussian distribution, into a complex distribution that approximates the given data distribution.
In particular, the process of DPMs involves iteratively adding noise to the clean image $\mathbf{x}0$ by a noise schedule $\beta{1:T}$. 
Here, $T$ denotes the total time step, and when $T$ is large enough, $\mathbf{x}_T$ is pure Gaussian noise.
Parameterized by a Markov chain, the process of adding noise to the clean images is referred to as the diffusion process.
\begin{equation}
\begin{aligned}
& q\left(\mathbf{x}_t \mid \mathbf{x}_{t-1}\right)=\mathcal{N}\left(\mathbf{x}_t ; \sqrt{1-\beta_t} \mathbf{x}_{t-1}, \beta_t \mathbf{I}\right), \\
& q\left(\mathbf{x}_{1: T} \mid \mathbf{x}_0\right)=\prod_{t=1}^T q\left(\mathbf{x}_t \mid \mathbf{x}_{t-1}\right).
\end{aligned}
\end{equation}
Diffusion Probabilistic Models (DPMs) are trained to recover the original image $\mathbf{x}_0$ from Gaussian noise $\mathbf{x}_T$ by gradually modeling the reverse of the transition distribution $q(\mathbf{x}_{t-1}\mid\mathbf{x}_{t})$.
However, calculating the posterior $q(\mathbf{x}_{t-1}\mid\mathbf{x}_{t})$ directly from $\mathbf{x}_{t}$ is a non-trivial task, and thus, DPMs are optimized in a maximum likelihood manner, akin to Variational AutoEncoders (VAEs).
\begin{equation}
\mathcal{L}_{DPM}=\mathbb{E}_{t \sim[1, T]} \mathbb{E}_{x_0 \sim p\left(x_0\right)} \mathbb{E}_{z_t \sim \mathcal{N}(0, \mathbf{I})}\left\|z_t-z_\theta\left(x_t, t\right)\right\|^2,
\end{equation}
where $z_\theta\left(x_t, t\right)$ is the training network predicting the noise in $\mathbf{x}_{t}$.
Benefiting from the intractable Markov chain and the simple loss function, DPMs provide satisfactory coverage of data distribution and can produce high-quality samples with sharp details and textures.
Furthermore, large-scale text-to-image diffusion models, such as Stable Diffusion~\cite{latentdiffusion} and DALLE-2~\cite{dalle2}, have empowered various downstream applications, including image-editing~\cite{ruiz2023dreambooth, gal2022image}, image-inpainting~\cite{lugmayr2022repaint, saharia2022palette, xie2023smartbrush, li2022sdm}, and so on.
Meanwhile, the development of data-efficient DPMs for limited-data generation has also garnered significant attention from the community~\cite{giannone2022few, wang2023patch, xia2023diffir}.

\noindent \textbf{Normalizing Flows (NFs).}
Normalizing Flows (NFs) learn a sequence of invertible transformations capable of mapping samples from a simple distribution to samples from a complex distribution.
Once trained, these transformations can be composed to form a complex function that captures the underlying structure of the original data.
To be more specific, NFs begin with a simple distribution, \emph{e.g.,} Normal distribution, and a series of invertible functions $f_{1: N}(\cdot)$ transform the simple distribution to the complex data distribution:
\begin{equation}
\mathbf{z}_i=f_{i-1}\left(\mathbf{z}_{i-1}\right).
\end{equation}
Note that $f_i$ is invertible thus the probability distribution of $z_i$ can be estimated by:
\begin{equation}
\begin{aligned}
p\left(\mathbf{z}_i\right) & =p\left(\mathbf{z}_{i-1}\right)\left|\frac{d f_i}{d \mathbf{z}_{i-1}}\right|^{-1}, \\
\log p\left(\mathbf{z}_i\right) & =\log p\left(\mathbf{z}_{i-1}\right)-\log \left|\frac{d f_i}{d \mathbf{z}_{i-1}}\right|.
\end{aligned}
\end{equation}
In this way, the final probability distribution is calculated by:
\begin{equation}
\log p\left(\mathbf{z}_N\right)=\log p\left(\mathbf{z}_0\right)-\sum_1^N \log \left|\frac{d f_i}{d \mathbf{z}_{i-1}}\right|.
\end{equation}
As evident from the learning scheme, NFs can generate samples from the target distribution exactly, rather than approximating it through sampling. 
However, their computational cost can be prohibitively high for large datasets and complex distributions.
Additionally, NFs struggle with high-dimensional data because of the curse of dimensionality.
The invertibility requirement further limits the synthesis performance, as the complexity of the transformations grows exponentially with the dimensionality of the data.

\begin{figure}[t]
    \centering
    \includegraphics[width=0.48\textwidth]{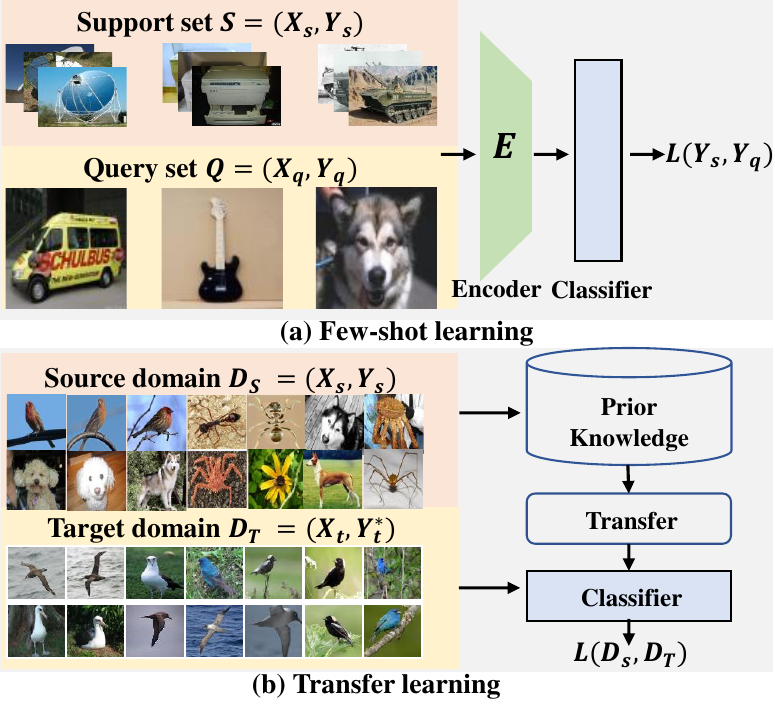}
    \caption{The general concepts of few-shot learning and transfer learning.}
    \label{fig:fewshot_transfer_learning}
\end{figure}

\subsection{Few-shot Learning}
Inspired by human's ability to learn new concepts from a few observations, few-shot learning (FSL), which seeks to learn novel classes from a \emph{few} samples, has gained significant attention.
Concretely, as demonstrated in~\cref{fig:fewshot_transfer_learning}, a dataset is divided into two sets of classes: \emph{base classes} $\mathbf{C}_b$ and \emph{novel classes} $\mathbf{C}_n$, where $\mathbf{C}_b \cap \mathbf{C}_n = \emptyset$.
Then, the model is trained on the \emph{base classes} $\mathbf{C}_b$ in an episodic task-by-task manner, 
where each episode consists of a \emph{support set} $\mathcal{S} = \left\{\left(\boldsymbol{x}_i, y_i\right)\right\}_{i=1}^{|\mathcal{S}|}$
and a \emph{query set} 
$\mathcal{Q}=\left\{\left(\boldsymbol{x}_j\right)\right\}_{j=1}^{|\mathcal{Q}|}$.
Finally, the model is expected to adapt the learned knowledge from the \emph{base classes} to the\emph{novel classes} $\mathbf{C}_n$, \emph{i.e.,} capturing samples of new categories efficiently from only a few samples.
This learning paradigm has made remarkable progress in various tasks of the few-shot learning field, such as classification~\cite{chi2021learning, chi2022better, yang2022few, 9634045, 9930675}, object detection~\cite{fan2020few, sun2021fsce} and segmentation~\cite{9839487, 9154595, liu2020crnet, li2021adaptive, lang2022learning}, as well as image generation~\cite{hong2020deltagan, hong2020f2gan, wavegan, ding2022attribute}.
Existing surveys on few-shot learning~\cite{wang2020generalizing, song2023comprehensive, antonelli2022few} mainly discuss classification and perception tasks, while generation remains largely unexplored. 
Therefore, a comprehensive review of few-shot generation is essential and complementary to prior work.
%
%

\subsection{Transfer Learning}
The generalization ability of traditional models is often impeded by the discrepancies between different domains. 
To address this challenge, transfer learning has been developed to explicitly reduce the shift across data distributions.
Transfer learning aims to transfer learned prior knowledge from a \emph{source domain} $\mathcal{D}_S$, 
where sufficient training data is available, to a \emph{target domain} $\mathcal{D}_T$ with limited data, as illustrated in~\cref{fig:fewshot_transfer_learning} (b).
As a mature and well-defined topic, transfer learning has been successfully applied in many practical domains, ranging from natural language processing (NLP)~\cite{houlsby2019parameter}, speech recognition~\cite{kunze2017transfer, xiong2020source, luo2021cross}, to various computer vision tasks~\cite{luo2021cross, 9512429, 9219132, wang2019class, bu2021gaia, xiao2022transfer}.
Based on the relationship between the source and the target domains, transfer learning can be further classified into several different types~\cite{pan2009survey}, such as multi-task learning, domain adaptation, unsupervised transfer learning, and so on.
In the field of image synthesis under limited data, transfer learning techniques are leveraged to reuse the pre-trained source-domain models and transfer relevant knowledge to improve the synthesis quality of the target domain, which will be elaborated in the following context.

\section{Data Efficient Generative Models}
\label{sec:data_efficient_generative_models}
In this section, we provide a detailed discussion and analysis of data efficient generative models. 
The problem of data-efficient generative modeling is defined in~\cref{sec:data_efficient_gen_definition}, followed by a summary of existing models categorized into four distinct types in~\cref{sec:data_efficient_gen_taxonomy}. 
Finally, we discuss popular benchmarks and the performance of existing models in~\cref{sec:data_efficient_benchmarks_performances}. 

\subsection{Problem Definition}
\label{sec:data_efficient_gen_definition}
Data efficient generative models refer to the scenario where a generative model is trained on a limited amount of training data, 
such as 100 images from 100-shot datasets~\cite{liu2020towards, DiffAug} or 1316 images from the MetFace~\cite{stylegan2-ada}, to produce diverse and plausible images that follow the given distribution.
However, several issues such as model overfitting and memorization are prone to occur when training a generative model on limited samples $D$ from scratch.
%
%
Additionally, the imbalance between the discrete limited training samples and continuous latent distribution might lead to unstable training~\cite{InsGen, li2022fakeclr, li2022comprehensive}.
Therefore, data efficient generative models are expected to have satisfactory data efficiency to capture the given distribution.
%
%
Notably, in contrast to the few-shot generative adaptation in~\cref{sec:few_shot_generative_adaptation}, data efficient generative models do not require pre-trained models available for further fine-tuning.
Lots of efforts have been endowed to improve the synthesis quality of data efficient generative models, and their advancements are discussed below.

\subsection{Model Taxonomy}
\label{sec:data_efficient_gen_taxonomy}
According to different techniques and intuition in data-efficient generative models, existing approaches could be categorized into four distinct categories.
Namely, augmentation-based, regularization-based, architecture variants, and off-the-shelf models-based approaches.
We will provide technical details of each approach below.

\noindent \textbf{Augmentation-based approaches.}
One straightforward solution for training generative models under limited data is to enlarge the training sets with data augmentation.
Various data augmentation techniques have been successfully applied to increase data diversity, thereby mitigating the overfitting of generative models under limited data~\cite{stylegan2-ada, jiang2021deceive, DiffAug}.
For instance, Karras \etal developed an adaptive discriminator augmentation (ADA) strategy to adaptively control the strength of data augmentation~\cite{stylegan2-ada}, 
Similarly, Zhao \etal proposed differentiable augmentation (DiffAug) to impose various types of differentiable augmentations on both real and fake samples~\cite{DiffAug}.
Furthermore, Jiang \etal designed adaptive pseudo augmentation (APA) to alleviate the overfitting by adaptively augmenting the real data with the generator itself, enabling healthier competition between the generator and the discriminator~\cite{jiang2021deceive}.
Formally, the objective of augmentation-based models is given as:
\begin{equation}
\begin{aligned}
\mathcal{L}_D & =-\mathbb{E}_{\mathbf{x} \sim p_{\mathbf{x}}}[\log (D(\mathbf{T}(\mathbf{x})))]-\mathbb{E}_{\mathbf{z} \sim p_{\mathbf{z}}}[\log (1-D(\mathbf{T}(G(\mathbf{z}))))], \\
\mathcal{L}_G & =-\mathbb{E}_{\mathbf{z} \sim p_{\mathbf{z}}}[\log (D(\mathbf{T}(G(\mathbf{z}))))],
\end{aligned}
\end{equation}
where $\mathbf{T}(\cdot)$ denotes various data augmentations.
In order to enable the generative model to work with more robust augmentations, Jeong proposed ContraD~\cite{jeong2021training}, which leveraged contrastive learning~\cite{bachman2019learning, oord2018representation, chen2020simple} to incorporate a wide range of data augmentations in GAN training.
Specifically, ContraD employed one network to extract a contrastive representation from a given set of data augmentations and samples (both real and generated images).
Then, the actual discriminator is defined upon the contrastive representation to minimize the training loss.
Notably, classical data augmentations, such as rotation and translation, have been identified as potentially manipulating the real distribution and misleading the generator to learn the distribution of the augmented data~\cite{tran2021data}.
Consequently, prior studies employ either differentiable~\cite{stylegan2-ada, DiffAug} or invertible~\cite{tran2021data} transformations to augment the training sets.
Moreover, Huang \etal randomly masked out spatial and spectral information of input images to encourage more challenging holistic learning from limited data~\cite{huang2022masked}.
The masked images can be viewed as augmenting images with random masks, enlarging the training set, and simultaneously building a robust discriminator.
Furthermore, besides applying augmentations at the image level, Dai \etal developed an implicit augmentation method to accomplish sample interpolation at the feature level~\cite{dai2022adaptive}.
In this way, the interpolated features could be viewed as new samples at the real data manifold, thus facilitating model training in low-data regimes.
Data augmentation is an intuitive and promising solution to improve data efficiency, and it is often complementary to other alternatives since no modification of the model and loss functions is performed.

\noindent \textbf{Regularization-based approaches.}
Regularization is a popular technique to stabilize the training of deep models by penalizing the training process with additional constraints~\cite{li2023systematic}.
In particular, various regularization approaches have been proposed~\cite{fang2022diggan, zhang2019consistency, zhao2021improved, kim2022feature} to mitigate the overfitting of data efficient generative models.
For instance, Tseng \etal first tracked the discriminator predictions with exponential moving average variables (\emph{i.e.,} anchors) and then calculated their proposed regularization term to push the discriminator to mix the predictions of real and generated images, enabling a more robust training objective~\cite{tseng2021regularizing}.
Similarly, Fang \etal regularized the discriminator by narrowing the gap between the norm of the gradients of the discriminator's prediction \emph{w.r.t} real images and \emph{w.r.t} generated images, thus avoiding bad attractors within the loss landscape~\cite{fang2022diggan}.
Furthermore, Yang developed a prototype-based regularization to improve fidelity and a variance-based regularization to facilitate diversity~\cite{yang2023protogan}.
In general, the regularization technique is orthogonal to other solutions since the network architecture remains unchanged.
For example, Zhao \etal augmented the real data with various augmentations and penalized the sensitivity of the discriminator to these augmentations with consistency regularization (CR)~\cite{zhang2019consistency}.
However, CR might introduce artifacts into the GAN samples since the augmentations are only applied to the real images, leading to an imbalanced training process.
To address this, Zhao \etal proposed balanced consistency regularization (bCR) to apply the augmentations on both real and generated samples.
They also introduced latent consistency regularization (zCR) to modulate the sensitivity of the generator and discriminator changes in the latent space~\cite{zhao2021improved}.
Following the same motivation of improving the discriminator, Kim \etal devised feature statistics mixing regularization (FSMR) to encourage the discriminator to be invariant to different styles of images~\cite{kim2022feature}.
This is accomplished by mixing features of an original image and a reference image in the feature space, generating images with novel styles in the semantic feature space.
However, the aforementioned methods still learned the discriminator by encouraging it to distinguish real and generated samples, which might provide insufficient feedback to the generator.
To combat this, Yang \etal assigned an additional instance discrimination task to the discriminator, which required the discriminator to distinguish every individual instance~\cite{InsGen}.
Considering that the synthesized images can be infinitely sampled, this approach provided the discriminator with sufficient samples to improve its representation ability.
In turn, the generator received meaningful feedback from the discriminator, enabling it to produce diverse images.

In addition to overfitting, generative models under limited data might display another undesirable property named latent discontinuity~\cite{li2022fakeclr}, which refers to the model that yields discontinuous transitions in the output space when smoothly interpolated in the latent space.
To address this, Kong \etal proposed a tow-side mixup-based distance regularization~\cite{kong2022few}.
They first sampled a random interpolation coefficient $\mathbf{c}$ from a Dirichlet distribution to enforce relative semantic distances between synthesized samples to follow the mixup ratio. 
Simultaneously, controlled interpolation on the discriminator's feature space enables the semantic mixups of scarce data points to be obtained and exploited to guide the feature space to align with semantic distances.
By doing so, both the latent space and feature space become smoother, and the latent space further enjoys mode-preserved properties.
In contrast, Yang \etal introduced a noise perturbation strategy to enforce the discriminator's invariance to small perturbations in the latent space, thus improving the discriminative power~\cite{InsGen}.
Following this philosophy, Li \etal revisited the noise perturbation scheme and devised three additional techniques, namely noise-related latent augmentation, diversity-aware queue, and forgetting factor of the queue,  to integrate contrastive learning into data efficient GAN training~\cite{li2022fakeclr}.
In particular, noise-related latent augmentation adopts different latent sampling to provide stronger similarity priors in low-density regions. 
Diversity-aware queue defines a dynamic queue size of negative samples based on the estimated sample diversity.
Moreover, the forgetting factor of the queue assigns higher importance to current synthesized samples and lower attention to previous synthesized samples in the negative queue.
Typically, regularization technique is a simple yet effective approach that introduces additional constraints or extra priors to improve model stability.
The technique is easy to implement since it requires no modifications to the network architecture or original loss functions.
Hence, regularization is often complementary to other solutions.
However, in limited-data regimes, stronger regularization is often required due to challenges such as model overfitting and memorization caused by data scarcity. 
Accordingly, finding more representative priors and additional supervision signals is crucial in developing more effective regularization techniques.

\noindent \textbf{Architecture variants.}
Another popular technique for improving the data efficiency of generative models is to design suitable network architectures. 
While style-based generative models~\cite{stylegan, stylegan2, stylegan3} have achieved impressive performance, they still struggle when given limited data due to their massive parameters, which lead to severe overfitting.
To address this issue, Liu \etal proposed a light-weight GAN, \emph{i.e.,} FastGAN~\cite{liu2020towards}, which constituted of a skip-layer channel-wise excitation module and a self-supervised discriminator to accomplish high-quality synthesis with minimum computing cost.
Building on the success of FastGAN, Li \etal proposed a memory concept attention (MoCA)~\cite{li2022prototype} to dynamically update and cache the prototype memories with a momentum encoder.
This attention mechanism can also modulate the continuous response of intermediate layers, allowing for a hierarchical and flexible composition of novel concepts to produce new images.
Yang \etal further enhanced the performance of FastGAN and MoCA by introducing a frequency-aware discriminator and a high-frequency alignment module, which aims to mitigate the unhealthy competition between the generator and the discriminator~\cite{wang2022fregan}.
In order to discover the optimal model designs of FastGAN, Shi \etal proposed AutoInfoGAN~\cite{shi2023autoinfo}, which leveraged a reinforcement learning neural architecture search (NAS) approach to identify the best network architectures.
Additionally, a contrastive loss was assigned to the discriminator to improve its representative ability.
Alternatively, Cui \etal proposed a generative co-training framework named GenCo~\cite{cui2022genco}, which incorporated multiple complementary discriminators into the model.
This framework enabled the generator to obtain diverse supervision from various distinct views of multiple discriminators.
However, the computational cost of GenCo was substantially higher than that of FastGAN since multiple independent discriminators were jointly trained.

Besides designing novel architectures, reducing the parameter complexity of existing large-scale models is also a promising solution for improving data efficiency.
For instance, Chen \etal discovered independently trainable and highly sparse subnetworks (\emph{i.e.,} lottery tickets~\cite{frankle2018lottery, zhou2019deconstructing}) from the original model and focused on learning the sparse subnetworks to enable a data-efficient generative model~\cite{chen2021data}.
However, finding GAN tickets required an additional resource-consuming process of train-prune-retrain, which is expensive in practice.
To address this, Saxena \etal proposed Re-GAN~\cite{saxena2023re}, which dynamically reconfigured the model architecture during training.
Specifically, Re-GAN repeatedly pruned unimportant connections of the model and grew the connections during training to reduce the risk of pruning important connections.
Considering that the discriminator overfits easily to limited data in the early stage of training, Yang \etal proposed DynamicD~\cite{yang2022improving} to gradually decrease the capacity of the discriminator.
Concretely, DynamicD randomly Shrank the layer width with a shrinking coefficient throughout the training process.
This scheme enabled decreased model capacity and introduced multiple discriminators benefiting from the random sampling.
Compressing the model capacity relaxes the data requirement since the pruned model is more light-weighted, and this solution is orthogonal to approaches that keep the model unchanged for a further performance boost.

\noindent \textbf{Off-the-shelf models.}
In contrast to generative modeling that is trained from scratch in an unsupervised manner, visual recognition tasks typically utilize off-the-shelf large-scale pre-trained models for downstream applications. 
These models have proven effective at capturing useful representations, thus one would naturally wonder whether these models can be employed in training generative models.
Sauer \etal made the first attempt to introduce pre-trained models into GAN training via projecting generated and real samples into a pre-defined feature space~\cite{projectedGAN}.
The projected features were then mixed across channels and resolutions to exploit the full potential of pre-trained perceptual feature spaces. 
This method significantly reduced the sample efficiency, convergence speed, and training time since only a small number of parameters were learned.
Mangla \etal transferred informative prior knowledge derived from self-supervised/supervised pre-trained networks to facilitate the GAN training~\cite{mangla2022data}.
In particular, they used representations of each instance of training data obtained from the pre-trained models as a prior instead of a data distribution itself.
However, with so many off-the-shelf models available in the community, it was still unclear which one(s) should be selected and in what manner they could be most effective.
Accordingly, Kumari \etal proposed to integrate the most accurate model by probing the linear separability between real and synthesized samples in the pre-trained model embedding space, enabling ensembled discriminators~\cite{kumari2022ensembling}.
It turned out that ensembling off-the-shelf models can improve GAN training in both limited data and large-scale settings.
Especially, by leveraging powerful pre-trained neural networks and a progressive growing strategy, StyleGAN-XL~\cite{styleganxl} achieved a new state-of-the-art performance on large-scale image synthesis tasks, \emph{e.g.,} obtaining the best $2.30$ FID score on $256 \times 256$ ImageNet~\cite{ImageNet}.

Unlike the above approaches that directly employed off-the-shelf models, Cui \etal proposed a knowledge distillation approach to leverage the prior information of pre-trained vision-language models~\cite{cui2023kd}.
Through distilling general knowledge from text-image paired information of the vision-language model (\emph{i.e.,} CLIP~\cite{radford2021learning}), the data efficiency was effectively improved.
Discriminative and generative models share similar objectives in learning meaningful representations of observed data. 
Therefore, stronger pre-trained visual perceptual models might bring further improvements to visual synthesis tasks, especially when training samples are limited. 
However, the selection of pre-trained models should be carefully considered since some models (e.g., Inception-V3~\cite{Inception}) might have a large perceptual null space~\cite{kynkaanniemi2022role}, leading to no actual performance gains~\cite{yang2023revisiting}.

\subsection{Benchmarks and Performances}
\label{sec:data_efficient_benchmarks_performances}
In this part, popular benchmarks for evaluating data efficient generative models are introduced,
and the performances of prior approaches on these benchmarks are summarized for a more comprehensive presentation.

\noindent \textbf{FFHQ.} 
The full set of FFHQ contains 70$K$ human-face images~\cite{FFHQgithub}, and it is the most commonly used dataset in the community.
In order to evaluate the data efficiency of various generative models, several subsets are randomly sampled from the full set, such as 1$K$, 2$K$ images, and these images are usually resized to 256 $\times$ 256 $\times$ 3 resolution.
Notably, although trained on a subset of all available training images, the quantitative metrics (\emph{e.g.,} FID~\cite{fid}, KID~\cite{KID}, LPIPS~\cite{zhang2018unreasonable}) are evaluated between 50$K$ synthesized images and the full 70$K$ images.
~\cref{tab:performance_metric_FFHQ} presents the FID scores of existing models under various data volumes of the FFHQ dataset.
We could tell from these results that 
1) Augmentation-based approaches such as ADA~\cite{stylegan2-ada} exhibit satisfactory compatibility to regularization-based methods (\emph{e.g.,} InsGen~\cite{InsGen}, FakeCLR~\cite{li2022fakeclr});
2) Various techniques present different data-efficiency under different data volumes, for example, InsGen surpasses other alternatives on 10$K$ images, whereas FakeCLR is the best under 2$K$ and 5$K$ training images.
Additionally, the potential of architecture variants methods on other types of approaches is under-explored.
More research is needed to investigate their compatibility for a further performance boost.

\setlength{\tabcolsep}{2pt}
\begin{table}[h]
\caption{FID ($\downarrow$) scores of previous methods on different amounts of training images of the FFHQ dataset~\cite{FFHQgithub}. The results are quoted from the published papers.} 
\centering
\resizebox{\columnwidth}{!} {
\begin{tabular}{l|c|c|c|c|c|c}
\toprule
\multirow{2}{*}{Method}          & \multirow{2}{*}{Type}  & \multicolumn{5}{c}{FFHQ}    \cr   
\cmidrule(r){3-7}
& & 0.1$K$  & 1$K$ & 2$K$ & 5$K$ & 10$K$ \cr
\cmidrule(r){1-1} \cmidrule(r){2-2} \cmidrule(r){3-3} \cmidrule(r){4-4}  \cmidrule(r){5-5} \cmidrule(r){6-6} \cmidrule(r){7-7}
StyleGAN2~\cite{stylegan2}& -             			 & 179.21      & 100.16     & 54.30   & 49.68   & 30.73  \cr
ADA~\cite{stylegan2-ada}  & Augmentation             & 82.17       & 21.29      & 15.39   & 10.96   &  7.29 \cr
CR~\cite{zhang2019consistency} 		       & Regularization           & 179.66      & -          & 71.61   & -       &  23.02 \cr
bCR + ADA~\cite{zhao2021improved} 		  & Regularization           & -           & 22.61      & -       & 10.58   &  7.53 \cr
DiffAug~\cite{DiffAug} 		  & Augmentation            & 61.91       & 25.66      & 24.32   & 10.45   &  7.86 \cr
DISP~\cite{mangla2022data} 		      & Off-the-shelf           & -       & -      & 21.06   & -   &  - \cr
GenCo~\cite{cui2022genco} 			  & Architecture           & 148         & 65.31      & 47.32   & 27.96   &  - \cr
APA + ADA~\cite{jiang2021deceive} 			  & Augmentation           & 65.31       & 18.89      & 16.91   & 8.38    &  - \cr
LeCam~\cite{tseng2021regularizing}			  & Regularization           & -           & 63.16      & -       & 23.83   &  14.58 \cr
InsGen + ADA~\cite{InsGen}  	      & Regularization           & 53.93       & 19.58      & 11.92   & -       &  4.9 \cr
FakeCLR + ADA~\cite{li2022fakeclr} 		  & Regularization           & 42.56       & 15.92      & 9.9     & 7.25    &  - \cr
Re-GAN~\cite{saxena2023re} 			  & Architecture           & -           & 36.30      & -       & 19.13   &  - \cr
DynamicD~\cite{yang2022improving} 		  & Architecture           & 50.37       & -          & 23.47   & -       &  - \cr
\bottomrule
\end{tabular}
}
\label{tab:performance_metric_FFHQ}
\end{table}

\noindent \textbf{AFHQ and CIFAR-10 datasets.}
AFHQ~\cite{choi2020stargan} consists of three sub-categories, including cat, dog, and wild, each containing about 5$K$ training images with the resolution of 512 $\times$ 512 $\times$ 3.
The three subsets are usually trained and tested individually in an unsupervised manner to evaluate the data efficiency.
Moreover, the full set of CIFAR-10~\cite{krizhevsky2009learning} contains 60$K$ training images with the resolution of 32 $\times$ 32 $\times$ 3.
Common choices are randomly sampling 10\%, 20\%, and 100\% of the samples from the full set for evaluation.
~\cref{tab:performance_metric_AFHQ} and~\cref{tab:performance_metric_CIFAR10} respectively show the FID quantitative results of prior methods on the AFHQ and CIFAR-10 datasets.
From these results, we can see that different types of methods are complementary to each other. 
Consequently, combining these techniques could lead to further performance improvements.
\setlength{\tabcolsep}{2pt}
\begin{table}[h]
\caption{FID ($\downarrow$) scores of previous data-efficient generative models on the AFHQ dataset~\cite{starganv2}. The results are quoted from the published papers.} 
\centering
\begin{tabular}{l|c|c|c|c}
\toprule
\multirow{2}{*}{Method}          & \multirow{2}{*}{Type}  & \multicolumn{3}{c}{AFHQ}    \cr
\cmidrule(r){3-5}
& & Cat  & Dog  & Wild \cr
\cmidrule(r){1-1} \cmidrule(r){2-2} \cmidrule(r){3-3} \cmidrule(r){4-4}  \cmidrule(r){5-5}
StyleGAN2~\cite{stylegan2}          & -             & 36.02      & 23.08      & 11.07 \cr
StyleGAN-ADA~\cite{stylegan2-ada}   & Augmentation  & 23.34      & 14.53      & 8.75  \cr
DiffAug~\cite{DiffAug}                     & Augmentation  & 14.50      & 12.15      & 9.89  \cr
FSMR + ADA~\cite{kim2022feature}                  & Regularization& 5.71       & 11.76      & 3.24  \cr
FSMR + DiffAug~\cite{kim2022feature}              & Regularization& 6.29       & 14.55      & 4.28  \cr
GenCo~\cite{cui2022genco}                       & Architecture  & 28.08      & 16.57      & 8.83  \cr
GenCo + ADA~\cite{cui2022genco}                 & Architecture  & 18.10      & 12.61      & 7.98  \cr
LeCam + ADA~\cite{tseng2021regularizing}                 & Regularization& 6.56       & -          & 2.47  \cr
Vision-Aided + ADA~\cite{kumari2022ensembling}          & Off-the-shelf & 2.69       & 4.81       & 2.36  \cr
ProjectedGAN~\cite{projectedGAN}                & Off-the-shelf & 2.16       & 4.52       & 2.17  \cr
\bottomrule
\end{tabular}
\label{tab:performance_metric_AFHQ}
\end{table}

\setlength{\tabcolsep}{2pt}
\begin{table}[h]
\caption{FID ($\downarrow$) scores of previous data-efficient generative models on the CIFAR-10 dataset~\cite{krizhevsky2009learning}. The results are quoted from the published papers.} 
\centering
\begin{tabular}{l|c|c|c|c}
\toprule
\multirow{2}{*}{Method}          & \multirow{2}{*}{Type}  & \multicolumn{3}{c}{CIFAR-10}    \cr 
\cmidrule(r){3-5}
& & 10\%  & 20\%  & 100\% \cr
\cmidrule(r){1-1} \cmidrule(r){2-2} \cmidrule(r){3-3} \cmidrule(r){4-4}  \cmidrule(r){5-5}
StyleGAN2~\cite{stylegan2}          & -             & 5.13      & 19.37     & 3.48 \cr
ADA~\cite{stylegan2-ada}            & Augmentation  & 3.55      & 7.40      & 3.05 \cr
APA~\cite{jiang2021deceive}                         & Augmentation  & 4.88      & -         & -    \cr
FastGAN~\cite{liu2020towards}                     & Architecture  & 10.17     & 25.36     & 7.30 \cr
FreGAN~\cite{wang2022fregan}                      & Architecture  & 6.62      & 20.75     & 6.37 \cr
DynamicD~\cite{yang2022improving}                    & Architecture  & 5.41      & 16.00     & 3.34 \cr
ContraD~\cite{jeong2021training}                     & Regularization& 7.16      & 3.82      & 2.54 \cr
InsGen~\cite{InsGen}                & Regularization& 2.60      & 5.44      & 1.77 \cr
FSMR + ADA~\cite{kim2022feature}                  & Regularization& 11.76     & 5.71      & 3.24 \cr
\bottomrule
\end{tabular}
\label{tab:performance_metric_CIFAR10}
\end{table}

\noindent \textbf{Low-shot datasets.}
In addition to sampling subsets from large-scale datasets, there are also single-category low-shot datasets~\cite{DiffAug} used for evaluating data-efficient generative models, such as Animal-Faces-Cat and Animal-Faces-Dog, which respectively contain 160 and 389 images for training, and 100-shot-Obama, Grumpy\_Cat (GCat), and Panda with 100 images.
These datasets are used to evaluate whether generative models can capture the data distribution of a low-shot dataset and produce diverse and novel images.
As for evaluation, 50$K$ generated images represent the synthesized distribution and the training set is employed as the referenced distribution.
The quantitative results of existing models on these datasets are presented in~\cref{tab:performance_metric_low_data}.
Despite existing models showing acceptable data efficiency and training stability on these low-shot datasets, there is a risk of memorizing the training images since these datasets contain only simple objects (\emph{e.g.,} clear cat and dog faces), leading to favorable synthesis diversity. 
Thus, it is recommended to consider evaluating data efficiency on datasets that contain more diverse objects.
Moreover, evaluation metrics should consider aspects of synthesized images in terms of fidelity, diversity, and distributional discrepancy.
Additionally, there are also some other benchmarks used for evaluating data-efficient generative models, such as higher resolution datasets like MetFace~\cite{stylegan2-ada} (1336 images with 1024 $\times$ 1024 $\times$ 3) and BrecaHAD~\cite{aksac2019brecahad} (136 images with 512 $\times$ 512 $\times$ 3) used in StyleGAN-ADA~\cite{stylegan2-ada}, as well as small subsets of CIFAR-100 and ImageNet evaluated in~\cite{tseng2021regularizing, saxena2023re}.
Interested readers can refer to the original papers for more experimental details.

\setlength{\tabcolsep}{2pt}
\begin{table}[h]
\caption{FID ($\downarrow$) scores of previous data-efficient generative models on the low-shot datasets~\cite{DiffAug}. 
The results are quoted from the published papers.
}
\centering
\resizebox{\columnwidth}{!} {
\begin{tabular}{l|c|c|c|c|c|c}
\toprule
\multirow{2}{*}{Method}          & \multirow{2}{*}{Type}  & \multicolumn{2}{c|}{Animal Faces} &  \multicolumn{3}{c}{100-Shot} \cr   
\cmidrule(r){3-4} \cmidrule(r){5-7} 
& & Cat  &  Dog & Obama & GCat & Panda \cr
\cmidrule(r){1-1} \cmidrule(r){2-2} \cmidrule(r){3-3} \cmidrule(r){4-4}  \cmidrule(r){5-5} \cmidrule(r){6-6} \cmidrule(r){7-7}
StyleGAN2~\cite{stylegan2}  & -         			     & 69.84       & 129.90     & 80.45   & 48.63   &  34.07 \cr
ADA~\cite{stylegan2-ada}   	& Augmentation            	 & 42.40       & 58.47      & 47.09   & 27.21   &  12.13 \cr
DiffAug~\cite{DiffAug}             & Augmentation               & 42.44       & 58.85      & 46.87   & 27.08   &  12.06 \cr
LeCam + DA~\cite{tseng2021regularizing} 			& Regularization             & 34.18       & 54.88      & 33.16   & 24.93   &  10.16 \cr
APA + ADA~\cite{jiang2021deceive} 		    & Augmentation               & 42.60       & 81.16      & 42.97   & 28.10   &  19.21 \cr
GenCo~\cite{cui2022genco} 				& Architecture               & 30.89       & 49.63      & 32.21   & 17.79   &  9.49  \cr
Lottery Ticket + DiffAug~\cite{frankle2018lottery}      & Architecture               & 47.40       & 68.28      & 52.86   & 31.02   &  14.75  \cr
InsGen~\cite{InsGen} 				& Regularization             & 33.01       & 44.93      & 32.42   & 22.01   &  9.85  \cr
FakeCLR~\cite{li2022fakeclr} 			& Regularization             & 26.34       & 42.02      & 26.95   & 19.56   &  8.42  \cr
FastGAN + DiffAug~\cite{liu2020towards} 	& Architecture               & 35.11       & 50.66      & 41.05   & 26.65   & 10.03  \cr
MoCA + DiffAug~\cite{li2022prototype} & Architecture               & 38.04       & 54.04      & 34.13   & 24.78   & 11.24  \cr
ProtoGAN + DiffAug~\cite{yang2023protogan}	& Architecture               & 33.31       & 50.89      & 35.46   & 25.29   & 9.52   \cr
FreGAN + DiffAug~\cite{wang2022fregan} 	& Architecture               & 31.05       & 47.85      & 33.39   & 24.93   & 8.97   \cr
Re-GAN~\cite{saxena2023re} 				& Architecture               & 42.11       & 57.20      & 45.70   & 27.36   & 12.60  \cr
AutoInfo-GAN~\cite{shi2023autoinfo}		& Architecture               & 33.33       & 49.72      & 35.54   & 24.83   & 9.36   \cr 
KD-GAN + ADA~\cite{cui2023kd}		& Off-the-shelf              & 32.81       & 51.12      & 31.78   & 19.76   & 8.85   \cr 
KD-GAN + LeCam~\cite{cui2023kd}		& Off-the-shelf              & 31.89       & 50.22      & 29.38   & 19.65   & 8.41   \cr 
\midrule
Scale/Shift~\cite{noguchi2019image} 		& Fine-tuning                & 54.83       & 83.04      & 50.72   & 34.20   & 21.38  \cr
MineGAN~\cite{wang2020minegan} 			& Extra module               & 54.45       & 93.03      & 50.63   & 34.54   & 14.84  \cr
TransferGAN~\cite{wang2018transferring} 		& Fine-tuning                & 52.61       & 82.38      & 48.73   & 34.06   & 23.20  \cr
TansferGAN+DA~\cite{wang2018transferring} 		& Fine-tuning                & 49.10       & 65.57      & 39.85   & 29.77   & 17.12  \cr
FreezeD~\cite{freezeD} 			& Fine-tuning                & 47.70       & 70.46      & 41.87   & 31.22   & 17.95  \cr
\bottomrule
\end{tabular}
}
\label{tab:performance_metric_low_data}
\end{table}

Although the approaches mentioned above have achieved impressive data efficiency trained from scratch with limited data, these models still tend to replicate the training images and produce less diverse outputs due to memorization. 
Accordingly, stronger and more effective techniques such as stronger regularization with extra supervision signals and novel network architectures are critical for further data efficiency improvements.

\section{Few-shot Generative Adaptation}
\label{sec:few_shot_generative_adaptation}
This section discusses the task of few-shot generative adaptation, which aims to transfer knowledge from a pre-trained large-scale source-domain generative model to a target domain with limited data.
In particular, the problem definition, a taxonomy of existing approaches, commonly used benchmarks and the performance are respectively provided in~\cref{sec:definision_few_shot_generative_adaptation}, ~\cref{sec:taxonomy_few_shot_generative_adaptation}, and~\cref{sec:benchmark_few_shot_generative_adaptation}.

\subsection{Problem Definition}
\label{sec:definision_few_shot_generative_adaptation}
~\cref{fig:problem_definition} illustrates the overall pipeline of few-shot generative adaptation.
Akin to transfer learning, the goal of few-shot generative adaptation is to transfer the knowledge of pre-trained generative models from large-scale source domains (\emph{e.g.}, FFHQ) to target domains with limited data (\emph{e.g.}, 10-shot images of baby faces) in a fast and efficient manner. 
Ideally, the adapted generative model should 
1) inherent the attributes of the source generative models that are invariant to the distribution shift, such as the overall structure, synthesis diversity, and semantic variances of generated images,
and
2) capture the internal distribution of the target domain to synthesize novel samples following the target distribution.
However, the limited amount of training data available for adaptation may cause the model to overfit, leading to model degradation.
Additionally, when the domain gaps between the source domain and the target domain are significant, negative transfer may occur, resulting in unrealistic generation.
Furthermore, inappropriate knowledge transfer~\cite{zhao2023exploring} may also lead to a deterioration in synthesis performance.
Below we present the primary solutions and discuss their characteristics regarding few-shot generative adaptation.

\subsection{Model Taxonomy}
\label{sec:taxonomy_few_shot_generative_adaptation}
The central idea of existing few-shot generative adaptation approaches is to preserve the useful knowledge of the source domain and adapt it to the target domain with limited data.
According to their techniques and design philosophy, we categorize prior studies of this field into four categories, namely 1) fine-tuning the model parameters to fit the target domain, 2) introducing extra branches to capture the target distribution, 3) regularizing the learning process via additional criteria, and 4) modulating the kernel of the network to transfer adequate knowledge.
In the following, we will introduce these methods in more detail and discuss their advantages and limitations.

\noindent \textbf{Fine-tuning.}
One typical solution for knowledge transfer in few-shot generative adaptation is to fine-tune the pre-trained generative model with the limited data of the target domain.
TransferGAN~\cite{wang2018transferring} was the initial attempt to transfer a pre-trained GAN by simply optimizing all parameter of the source model with the original GAN loss (see ~\cref{eq:GAN_equation}).
However, this straightforward strategy may lead to overfitting, especially when the target data are extremely limited. (\emph{e.g.,} 10-shot samples).
To address the overfitting issue, FreezeD~\cite{freezeD} fixed some low-level layers of the discriminator during the adaptation process.
Moreover, Zhao \etal revealed that low-level filters of both the generator and the discriminator can be transferred to facilitate more diverse generation for the target domain~\cite{zhao2020leveraging}.
An adaptive filter modulation was further developed to better adapt the filters of adaptation, enabling boosted diversity.
Another approach is to update only partial parameters of the pre-trained model to reduce overfitting.
For instance, Noguchi \etal proposed batch statistics adaptation (BSA)~\cite{noguchi2019image}, which focused on updating the parameters for batch statistics, scale and shift, of the generator's hidden layers, reducing the amount of parameters while maintaining the synthesis quality.
Similarly, elastic weight consolidation (EWC)~\cite{li2020few} identified the importance of model parameters and penalized the change of important parameters.
Furthermore, Robb \etal repurposed component analysis techniques for generative adaptation~\cite{robb2020few} by learning to adapt the singular values of the pre-trained weights with the corresponding singular vectors frozen. 
These methods constrain the changes of the parameters and reduce overfitting.
However, these methods focus solely on fitting the target distribution and may lose the prior knowledge of the source domain that cannot be derived from the limited data of the target domain. 
Accordingly, these methods may not fully exploit the potential of the pre-trained model and may not generate diverse and high-quality outputs.

\noindent \textbf{Extra branches.}
In order to identify the most beneficial knowledge to transfer for a specific target domain, Wang \etal proposed MineGAN~\cite{wang2020minegan}, which employed an additional mining network to find the distributions of pre-trained GANs that produce samples closest to the target images.
This mining network shifted the input distribution towards the most interested regions regarding the target distribution, enabling more effective and efficient knowledge transfer.
Similarly, Yang \etal imported two extra lightweight modules for generative adaptation~\cite{yang2021one}.
The first module is an attribute adaptor on the latent code to transfer the most distinguishable characters, while the second module is an attribute classifier attached to the discriminator to encourage the generator to capture appropriate characters from the target domain. 
These two modules are fast to optimize and bring appealing results under various settings, especially when only one reference image is available.
Along this line, Wu \etal proposed a domain re-modulation (DoRM) structure~\cite{wu2023domain}, which incorporated new mapping and affine modules to capture the characteristics of the target domain. 
DoRM also enabled multi-domain and hybrid-domain adaptation by integrating multiple mapping and affine modules.

With the remarkable development of diffusion models, it is interesting to investigate their performance for knowledge transfer to limited target domains.
Moon \etal demonstrated that fine-tuning only small subsets of pre-trained diffusion models' parameters can efficiently capture the target distribution~\cite{moon2022fine}.
They also proposed a time-aware adapter module to improve the synthesis quality by fitting inside the attention block of the pre-trained diffusion models according to different timesteps.
Zhu \etal designed a pairwise adaptation model to preserve useful information of the source domain by keeping the relative pairwise distances between synthesized samples~\cite{zhu2022few}. 
Such that, the diversity and synthesis details of original model were well-preserved, enabling diverse generation for the target domain.
Although introducing additional modules for generative adaptation is effective and efficient compared to fine-tuning-based approaches, their output images might resemble the source domain since the original generator remains unchanged. 
Therefore, it is important to strike a balance between utilizing the prior knowledge of the pre-trained model and adapting the model to the characteristics of the target domain to generate diverse and high-quality samples.

\noindent \textbf{Regularization.}
Another solution for knowledge transfer is to explicitly introduce additional supervision or constraint in the adaptation process.
For instance, Ojha \etal proposed two novel strategies to transfer the diversity information from a large-scale source domain to the target domain~\cite{ojha2021few}.
In particular, a cross-domain consistency (CDC) regularization item was integrated to preserve relative pairwise distances between the source and target generated images.
An anchor-based approach was further designed to encourage different levels of synthesis fidelity in the latent space to mitigate overfitting.
In order to align the spatial structural information between synthesized image pairs of the source and target, Xiao \etal developed a relaxed spatial structural alignment (RSSA)~\cite{xiao2022few}, which preserved and transferred the structural information and spatial variation tendency of the source domain to the target by compressing the latent space to a subspace close to the target domain and regularizing the self-correlation consistency and disturbance correlation consistency.
Following the same idea of preserving the diversity of the source domain, Hou \etal proposed a dynamic weighted semantic correspondence (DWSC) to explicitly preserve the perceptual semantic consistency between generated images of source and target domain~\cite{hou2022dynamic}.
Zhao \etal discovered that all generative adaptation models achieved similar quality after convergence, and thus proposed a dual contrastive learning (DCL) framework to slow down the diversity degradation by preserving the multi-level diversity of the source domain throughout the adaptation process.~\cite{zhao2022closer}.
For the global-level knowledge transfer, Zhang \etal leveraged the difference between the CLIP features of the source and target domain to constrain the target generator~\cite{zhang2022towardsoneshot}.
An attentive style loss that aligned the intermediate token between the adapted source image with the referenced target image was further integrated for local-level adaptation.
Moreover, Zhang \etal proposed a generalized generative adaptation framework that realized both style and entity transfer~\cite{NEURIPS2022_58ce6a4b}.
The core intuition behind this was to employ sliced Wasserstein distance to regularize the internal distribution of the referenced target images and the synthesized samples.
An auxiliary network was developed to explicitly disentangle the entity and style of referenced images, and a style fixation module was employed to obtain the exemplary style. 
A variational Laplacian regularization was further devised to improve the smoothness of the adapted generator.
Differently, Mondal \etal observed that target images can be `embedded' onto the latent space of a pre-trained model on source images~\cite{mondal2022few}.
Therefore, they optimized a latent learner network during the inference stage to find corresponding latent code to the target domain on the manifold of the source domain.
In this way, the target embedding is employed by the source-domain generator to produce novel images.
Following this, the most recent WeditGAN~\cite{duan2023weditgan} achieved knowledge transfer via relocating the distribution of source latent spaces towards target latent spaces by learning a constant latent offset for editing the latent space. 
Inspired by the remarkable generation capability of text-to-image diffusion models, Song \etal demonstrated that the generators can distill prior knowledge from large-scale text-to-image diffusion models by employing the classifier-free guidance as a critic~\cite{song2022diffusion}.
Additionally, a directional and reconstruction regularizer was developed to avoid model collapse.
This work revealed the potential of distilling prior knowledge from pre-trained large-scale diffusion models to other types of generative models.
Stronger regularization and further investigation on this topic are interesting future work.
Overall, regularization-based approaches render promising results for transferring knowledge from large-scale pre-trained generative models.
However, one critical limitation is that they have a trade-off between preserving source domain priors and modeling target distributions. 
That is, too-strong regularization leads to overfitting, and too-weak regularization causes source domain output instead of the target domain. 
Consequently, finding a proper value of regularization is critical to achieving effective and efficient knowledge transfer. 
Meanwhile, developing stronger regularization techniques that leverage more prior information/supervision signals of the original data is also essential for further improvements.

\noindent \textbf{Kernel modulation.}
One significant limitation of the aforementioned approaches is their sole reliance on the source domain,
which disregards the target domain/adaptation and raises concerns about the generalization capability of these models for setups with varying proximity between the source and target domains. 
To address this, Zhao \etal proposed an adaptation-aware kernel modulation (AdAM) pipeline~\cite{zhao2022few}.
In particular, AdAM probed the importance of different kernels in the network and preserved crucial weights during the adaptation process.
Building on this philosophy, they further devised RICK~\cite{zhao2023exploring}, which explored incompatible knowledge transfer in the adaptive process.
AdAM's method of estimating the importance of various filters was utilized. 
Next, filters with lower importance below a predefined threshold were pruned, and more important filters were frozen. 
The rest of the filters were fine-tuned in the training process.
This procedure successfully removed incompatible knowledge during target adaptation.
However, this technique may not be adequate for more challenging setups with a wide gap between the source and target domains, as crucial kernels/filters may be scarce for adaptation. 
Future research is needed to develop more effective approaches that consider both the source and target domains in the adaptation process to ensure the models' generalization capability.

\subsection{Benchmarks and Performances}
\label{sec:benchmark_few_shot_generative_adaptation}
In this part, we present popular benchmarks of few-shot generative adaptation, and compare the performances of prior studies on these benchmarks.

\noindent \textbf{FFHQ (source) to relevant human face target domains.}
Typically, the performance of few-shot generative adaptation approaches is evaluated under varying degrees of proximity between the source and target domains. 
The Babies, Sunglasses, and Sketches datasets~\cite{ojha2021few}, each containing only 10-shot target images for adaptation, are the most commonly used datasets for this purpose. 
Furthermore, the pre-trained FFHQ dataset, which comprises 70,000 training images, is the prevalent choice of source generator.
Quantitative evaluation entails computing various metrics on the entire set of Babies, Sunglasses, and Sketches datasets, consisting of approximately 2500, 2700, and 300 images, respectively.
~\cref{tab:performance_metric_FFHQ_adaptation_babies} presents the FID scores of existing alternatives on the three datasets.
The results indicate that:
1) augmentation-based approaches (\emph{e.g.,} ADA~\cite{stylegan2-ada}) are complementary to fine-tuning based few-shot generative adaptation models;
2) among all prior methods, regularization and the introduction of extra branches appear to be more effective than fine-tuning-based approaches, enabling better knowledge transfer from the source to the target.
However, the intersection of different types of methods remains underexplored, and it is essential to combine them to investigate the potential for further performance improvements.

\setlength{\tabcolsep}{2pt}
\begin{table}[h]
\caption{FID ($\downarrow$) scores of existing few-shot generative adaptation methods on transferring prior knowledge from the FFHQ source dataset to 10-shot target domains Babies, Sunglasses, and Sketches. 
The results are quoted from the published papers.
}
\centering
\resizebox{\columnwidth}{!} {
\begin{tabular}{l|c|c|c|c}
\toprule
\multirow{2}{*}{Method}          & \multirow{2}{*}{Type}  & \multicolumn{3}{c}{Source: FFHQ}    \cr    
\cmidrule(r){3-5}
& & Babies  & Sunglasses  & Sketches \cr
\cmidrule(r){1-1} \cmidrule(r){2-2} \cmidrule(r){3-3} \cmidrule(r){4-4}  \cmidrule(r){5-5}
TransferGAN~\cite{wang2018transferring}                 & fine-tuning             & 104.79      & 55.61      & 53.41 \cr
TransferGAN + ADA~\cite{wang2018transferring}  			& fine-tuning             & 102.58      & 53.64      & 66.99 \cr
Scale/Batch~\cite{noguchi2019image}  				& fine-tuning             & 140.34      & 76.12      & 69.32 \cr
FreezeD~\cite{freezeD}  					& fine-tuning             & 110.92      & 51.29      & 46.54 \cr
MineGAN~\cite{wang2020minegan}  					& extra-modules           & 98.23       & 68.91      & 64.34 \cr
EWC~\cite{li2020few}  						& fine-tuning             & 87.41       & 59.73      & 71.25 \cr
CDC~\cite{ojha2021few}  						& regularization          & 74.39       & 42.13      & 45.67 \cr
DWSC~\cite{hou2022dynamic}                        & regularization          & 73.37       & 36.04      & 39.86 \cr
DCL~\cite{zhao2022closer}  						& regularization          & 52.56       & 38.01      & 37.90 \cr
ADAM~\cite{zhao2022few}  						& modulation              & 48.83       & 28.03      & -	 \cr
RICK~\cite{zhao2023exploring}  						& modulation              & 39.39       & 25.22      & -     \cr
KD-GAN~\cite{cui2023kd}  					& modulation              & 68.67       & 34.61      & 35.87 \cr
GenDA~\cite{yang2021one}  						& extra-modules           & 47.05       & 22.62      & 31.97 \cr
WeditGAN~\cite{duan2023weditgan}                    & regularization          & 46.70       & 28.09      & 38.44 \cr
ISLL~\cite{mondal2022few}           & regularization          & 63.31       & 35.64      & 35.59 \cr
DoRM~\cite{wu2023domain}  		    & extra-modules           & 30.31       & 17.31      & 40.05 \cr
\bottomrule
\end{tabular}
}
\label{tab:performance_metric_FFHQ_adaptation_babies}
\end{table}

\noindent \textbf{FFHQ (source) to irrelevant animal face target domains.}
In addition to transferring knowledge from pre-trained source domains to relevant target domains, evaluating the model's performance under irrelevant source-target settings is crucial to determine its effectiveness.
Given a generator that is pre-trained on the human face domain (\emph{i.e.,} FFHQ), the generator is then adapted to the animal face target domains, namely AFHQ-Cat, AFHQ-Dog, and AFHQ-wild.
~\cref{tab:performance_metric_FFHQ_adaptation_AFHQ} shows the quantitative results under this setting.
Same conclusions akin to~\cref{tab:performance_metric_FFHQ_adaptation_babies} could be drawn from these results.
Interestingly, kernel modulation based methods (\emph{i.e.,} ADAM~\cite{zhao2022few} and RICK~\cite{zhao2023exploring}) outperform other alternatives with a substantial margin, demonstrating that mining important weights is effective for distant domain knowledge transfer.

\setlength{\tabcolsep}{2pt}
\begin{table}[h]
\caption{FID ($\downarrow$) scores of existing few-shot generative adaptation methods on transferring prior knowledge from the FFHQ source dataset to animal face target domains. 
The results are quoted from the published papers.
}
\centering
\resizebox{\columnwidth}{!} {
\begin{tabular}{l|c|c|c|c}
\toprule
\multirow{2}{*}{Method}          & \multirow{2}{*}{Type}  & \multicolumn{3}{c}{Source: FFHQ}
\cr    
\cmidrule(r){3-5}
& & AFHQ-Cat  & AFHQ-Dog  & AFHQ-Wild \cr
\cmidrule(r){1-1} \cmidrule(r){2-2} \cmidrule(r){3-3} \cmidrule(r){4-4}  \cmidrule(r){5-5}
TransferGAN~\cite{wang2018transferring}                 & fine-tuning             & 64.68      & 151.46      & 81.30 \cr
TransferGAN + ADA~\cite{wang2018transferring}  			& fine-tuning             & 80.16      & 162.63      & 81.55 \cr
FreezeD~\cite{freezeD}  					& fine-tuning             & 63.60      & 157.98      & 77.18 \cr
EWC~\cite{li2020few}  						& fine-tuning             & 74.61      & 158.78      & 92.83 \cr
CDC~\cite{ojha2021few}  						& regularization          & 176.21     & 170.95      & 135.13\cr
DCL~\cite{zhao2022closer}  						& regularization          & 156.82     & 171.42      & 115.93\cr
ADAM~\cite{zhao2022few}  						& modulation              & 58.07      & 100.91      & 36.87 \cr
RICK~\cite{zhao2023exploring}  						& modulation              & 53.27      & 98.71       & 33.02 \cr
\bottomrule
\end{tabular}
}
\label{tab:performance_metric_FFHQ_adaptation_AFHQ}
\end{table}

\noindent \textbf{Knowledge transfer between various source-target domains.}
In addition to evaluating the transfer of knowledge between different target domains, generators pre-trained on various source domains can also be used for knowledge transfer.
For instance, generative models trained on Cars (\emph{resp.} Church)~\cite{lsun} can be employed for other target domains, such as Abandoned Cars (\emph{resp.} Haunted House)~\cite{zhao2022few}.
Furthermore, Otto's Paintings dataset is utilized as the target domain for transferring prior information from the FFHQ source domain.
~\cref{tab:performance_metric_adaptation_other_domain} presents the LPIPS~\cite{zhang2018unreasonable} scores under these source-target domains.
Notably, the LPIPS score reflects the sample diversity of the target domain outputs.
We could observe that regularization-based methods perform better than fine-tuning-based models, suggesting the effectiveness of introducing extra supervision/prior information to the adaptation. 
By contrast, the performance of importing extra branches and modulating the kernels/filters of the network under such settings still remains vacant.
Thus, further investigation is required for future research.
Additionally, it is crucial to test the generalization capability of these models under more challenging scenarios, such as 
1) when only one single target image for adaptation~\cite{yang2020one, kwon2023one};
and
2) when multiple target domains are given, and source knowledge must be simultaneously adapted in a unified framework~\cite{kim2022dynagan, nitzan2023domain}

\setlength{\tabcolsep}{2pt}
\begin{table}[t]
\caption{LPIPS ($\uparrow$) scores of existing few-shot generative adaptation methods under distant source-target domain transfer.
The results are quoted from the published papers.
}
\centering
\resizebox{\columnwidth}{!} {
\begin{tabular}{l|c|c|c|c}
\toprule
\multirow{2}{*}{Method}          & \multirow{2}{*}{Type} & FFHQ$\rightarrow$ & Church$\rightarrow$ & Cars$\rightarrow$ \cr
 & & Otto's Paintings & Haunted House & Abandoned Cars \cr
\cmidrule(r){1-1} \cmidrule(r){2-2} \cmidrule(r){3-3} \cmidrule(r){4-4}  \cmidrule(r){5-5}
TransferGAN~\cite{wang2018transferring}                 & fine-tuning             & 0.51$\pm$0.02      & 0.52$\pm$0.04      & 0.46$\pm$0.03 \cr
TransferGAN + ADA~\cite{wang2018transferring}  			& fine-tuning             & 0.54$\pm$0.02      & 0.57$\pm$0.03      & 0.48$\pm$0.04 \cr
Scale/Shift~\cite{noguchi2019image}  			    & fine-tuning             & 0.46$\pm$0.02      & 0.43$\pm$0.02      & 0.41$\pm$0.03 \cr
FreezeD~\cite{freezeD}  					& fine-tuning             & 0.54$\pm$0.03      & 0.45$\pm$0.02      & 0.50$\pm$0.05 \cr
MineGAN~\cite{wang2020minegan}						& extra-modules           & 0.53$\pm$0.04      & 0.44$\pm$0.06      & 0.49$\pm$0.02 \cr
EWC~\cite{li2020few}  						& fine-tuning             & 0.56$\pm$0.03      & 0.58$\pm$0.06      & 0.43$\pm$0.02 \cr
CDC~\cite{ojha2021few}  						& regularization          & 0.63$\pm$0.03      & 0.60$\pm$0.04      & 0.52$\pm$0.04 \cr
DCL~\cite{zhao2022few}  						& regularization          & 0.66$\pm$0.02      & 0.63$\pm$0.01  	& 0.53$\pm$0.02 \cr
\bottomrule
\end{tabular}
}
\label{tab:performance_metric_adaptation_other_domain}
\end{table}

\section{Few-shot Image Generation}
\label{sec:few_shot_image_generation}
This section discusses the task of few-shot image generation, whose problem definition, main solutions, popular benchmarks and performances are respectively presented in ~\cref{sec:definition_few_shot_image_generation},~\cref{sec:taxonomy_few_shot_image_generation}, and~\cref{sec:benchmarks_few_shot_image_generation}.

\subsection{Problem Definition}
\label{sec:definition_few_shot_image_generation}
Following the prior philosophy of few-shot learning, few-shot image generation is formulated to synthesize diverse and photorealistic images for a new category given $K$ input images from the same category.
The model is trained in an episodic task-by-task manner, wherein each $N$-way-$K$-shot image generation task is defined by $K$ input images from each of the $N$ classes.
The training and testing phases of few-shot image generation involve splitting the dataset into two disjoint subsets: seen classes $\mathbb{C}_s$ and unseen classes $\mathbb{C}_u$.
During training, a considerable number of $N$-way-$K$-shot image generation tasks from $\mathbb{C}_s$ is randomly sampled, with the aim of encouraging the model to acquire the ability to generate novel samples. 
In the testing phase, the model is expected to generalize this ability to generate new images for $\mathbb{C}_u$, based on only a few samples from each class.
Few-shot image generation is known to suffer from catastrophic forgetting, whereby the model forgets previous knowledge and focuses excessively on new tasks, thus impairing its ability to generalize to unseen classes.
Exiting approaches seek to address this challenge from various perspectives, we analyze their respective advantages and disadvantages in detail below.
\subsection{Model Taxonomy}
\label{sec:taxonomy_few_shot_image_generation}
Depending on the mechanisms of producing novel images with few input conditional images, prior methods of few-shot image generation can be categorized into optimization-based, fusion-based, transformation-based, and inversion-based approaches.
These approaches differ in terms of model design, optimizing objectives, model training, and inference, we present their primary concepts in detail below.

\noindent \textbf{Optimization-based approaches.}
Motivated by the tremendous advances in meta-learning~\cite{finn2017model, 9428530}, particularly in the context of few-shot classification tasks~\cite{finn2018probabilistic, antoniou2018train, chen2021meta, jamal2019task}, optimization-based few-shot generation models have been proposed to leverage meta-learning for learning episodic tasks and generating novel samples from few observations.
To be more specific, these models aim to learn optimal parameters $\Phi$, via meta-training on a set of tasks $T$ with multiple tasks $\tau$, where each task $\tau$ defines an image generation problem with few conditional images $\mathbf{X}_{\tau}$ and a loss $L_{\tau}$ that measures the score of discriminating generated images from real images sampled from task $\mathbf{X}_{\tau}$.
The goal is to enable fast adaptation to novel random tasks by minimizing the associated loss $L_{\tau}$:
\begin{equation}
\operatorname{min}_{\Phi} \mathbb{E}_\tau\left[L_\tau\left(U_\tau^k(\Phi)\right)\right],
\end{equation}
where $U_\tau^k(\Phi)$ represents the operator that update the parameters $\Phi$ conditioned on the tasks $T$.
Several optimization-based few-shot generation models have been proposed in the literature. 
As a pioneering attempt, FIGR~\cite{clouatre2019figr} incorporated the optimization-based meta-learning Reptile~\cite{nichol2018first} into GAN training for few-shot image generation.
Similarly, Liang \etal proposed a plug-and-play domain adaptive few-shot generation framework (DAWSON) which supported a broad family of meta-learning models and various GANs~\cite{liang2020dawson}.
However, training these models is computationally expensive due to the requirement of two-stage training for both GANs and meta-learning.
To address this issue, Phaphuangwittayakul \etal developed a fast adaptive meta-learning (FAML) framework, which significantly reduced training time by training a simpler network with conditional feature vectors from the encoder~\cite{phaphuangwittayakul2021fast}.
Moreover, they extended FAML via applying a self-supervised contrastive learning strategy to the fast adaptive meta-learning framework~\cite{phaphuangwittayakul2022few}, which improved both the speed of model convergence and the synthesis performance.
Nevertheless, the output images produced by optimization-based models often suffer from poor fidelity and unrealistic appearance, despite their ability to produce novel samples.

\noindent \textbf{Transformation-based approaches.}
Transformation-based approaches aim to capture the inter and intra-category translation to produce new images of unseen categories.
For instance, DAGAN~\cite{antoniou2017data} produced novel samples by injecting random noises into the representation of a single input image.
However, conditioned on a single input image, the sample diversity of images produced by DAGAN~\cite{antoniou2017data} was limited.
To mitigate this, Hong \etal proposed a delta generative adversarial network (DeltaGAN)~\cite{hong2022deltagan}, which explicitly extracted the intra-category information (\emph{a.k.a} ``delta") from same-category feature pairs during training to produce novel features and transform input conditional images into new images, thereby substantially improving synthesis diversity.
In order to learn the relationship across seen and unseen categories, Huang \etal proposed an implicit support set autoencoder (ISSA)~\cite{huang2021few}, which inferred the representation of the underlying distribution from trained ISSA to produce novel samples.
However, the transformation-based methods might fail to produce novel images when the intra- and inter-category transformation relations are complex.
One solution is to build more advanced modules to encode the complex transformations between distinct samples.
Following this philosophy, Hong \etal proposed Disco-FUNIT~\cite{hong2022fewdisco}, a discrete vector quantization-based transformation method.
Concretely, they first learned a compact dictionary of local content vectors through quantizing continuous content mapping.
The encoded discrete content maps captured the translation relationships between various samples.
Then, the autoregressive distribution of the discrete content vectors was modeled conditioned on style maps, alleviating the incompatibility between content and style maps.
Finally, diverse images could be produced conditioned on the style vectors and the content maps.
Similarly, Xie \etal proposed FeaHa~\cite{xie2022learning}, which explicitly learned and memorized reusable features from seen categories to generate new features for novel image generation.
The image features were decomposed into category-related and category-independent features, the category-independent features were employed to generate new features with a feature hallucination module.
Through sampling from the memorized reusable features, diverse and novel images can be generated.
The successes of Disco-FUNIT~\cite{hong2022fewdisco} and FeaHa~\cite{xie2022learning} demonstrate that transformation-based approaches benefit from stronger representations of the conditional input images, such as semantic, content, and style features.
This suggests that future research in this area may benefit from exploring the development of more effective representation learning strategies.

\noindent \textbf{Fusion-based approaches}.
Fusion-based models interpolate the conditional input images at the image or semantic level to generate novel samples.
For instance, MatchingGAN~\cite{matchinggan} fused the features of input images from the same class with random vectors and generated new images for this class based on the fused features. 
GMN~\cite{bartunov2018few} combined VAE~\cite{vae} with Matching networks~\cite{vinyals2016matching} to capture the few-shot distribution.
Although these models can generate diverse images, the details of the synthesized samples may be unfavorable.
To address this issue, F2GAN~\cite{hong2020f2gan} employed a non-local attention module to explicitly attend to more low-level details, resulting in a better generation of image details.
In particular, after fusing the high-level features of conditional images with random interpolation coefficients, a non-local attention module was employed to explicitly attend more low-level details to the fused features, enabling better generation of image details.
However, MathingGAN and F2GAN suffer from poor semantic fine-grained details due to severe semantic misalignment caused by global perspective interpolation of conditional images.
To tackle this, Gu \etal proposed a local fusion GAN model (LoFGAN)~\cite{lofgan}, enabling local semantic fusion.
Concretely, the input conditional images were first randomly divided into a base image and several reference images, where semantic similarities between local representations of the base image and the reference images were computed with the cosine distance, and the closest relevant representations are fused with the base image. enabling highly correlated semantic fusion.
Together with a local reconstruction that facilitated the fidelity of fused local representations, the synthesis quality of LoFGAN surpassed prior studies by a substantial margin.
Based on LoFGAN, Li \etal further improved the synthesis performance with an adaptive multi-scale modulation GAN (AMMGAN)~\cite{li2023ammgan}, whose key contribution was introducing an adaptive self-metric fusion module that adaptively adjusted the mean and variance of the fused feature based on the high-level semantic features from the decoder.

F-principal~\cite{xu2019frequency} identified that deep neural networks preferentially capture frequency signals from low to high, with low frequency signals having higher priority than high frequency signals in the fitting process~\cite{gao2021high, schwarz2021frequency, jiang2021focal}.
This phenomenon also exists in deep generative models, where they prefer to produce low frequency signals with more superficial complexity.
In order to alleviate the generator's struggles of capturing high frequency components, Yang \etal designed a frequency-aware GAN dubbed WaveGAN~\cite{wavegan},
which decomposed features into multiple frequency representations from low to high and feed the high frequency components to the decoder via high-frequency skip connections, improving the frequency awareness of the generator.
Low frequency skip connections and a frequency loss were further employed to maintain the textures and outline of generated images.
Experimental results demonstrated that explicitly incorporating high-frequency signals into the generating process significantly improves the synthesis fidelity. 
However, it is still unclear which frequency components are the most important for high-fidelity few-shot image generation, and how to improve diversity with frequency signals.
Further studies are required to investigate these aspects of frequency-based fusion approaches for few-shot image generation.

\noindent \textbf{Inversion-based approaches.}
GAN inversion~\cite{inversionsurvey} aims to obtain the inverted latent code via inverting a given image back into the latent space of a pre-trained GAN model, which can be used for reconstructing~\cite{shen2020interfacegan, shen2020interpreting}, manipulating~\cite{yang2021semantic}, and editing~\cite{bai2022high} the original image.
Based on the successes of GAN inversion techniques, it is natural to investigate their potential impacts on few-shot image generation.
Based on the assumption that any image is a composition of attributes and the editing direction for a specific attribute should be the same regardless of different categories, inversion-based approaches were first developed in attribute group editing (AGE)~\cite{ding2022attribute} by Ding \etal.
Specifically, they employed class embeddings to represent the category-relevant attributes and learned a dictionary to collect the category-irrelevant attributes.
The class embeddings and the dictionary were learned in the latent space through inversion, so that diverse images could be generated by editing the category-irrelevant attributes in the latent space.
Moreover, through manipulating the latent codes toward specific directions of the attributes, this method enabled controllable image generation for the first time in this field.
In order to achieve more stable class-relevant image generation, Ding \etal further extended AGE with an adaptive editing strategy to enhance the stability of category retention~\cite{ding2023stable}.
However, Li \etal found that AGE suffered from the trade-off between the quality and diversity of generated images and thus proposed a hyperbolic attribute editing-based (HAE) method to capture the hierarchy among input conditional images~\cite{li2022euclidean}.
Their editing of the attributes was accomplished hierarchically in hyperbolic space, which can encode semantic representations with a large corpus of images, enabling a more controllable and interpretable editing process.
In contrast to the above methods, which viewed the attributes in the latent space as discrete properties, Zheng \etal explored the continuity of the latent space for finding unseen categories~\cite{zheng2023my}.
The rationale behind their method was that the neighboring latent space around the novel class belongs to the same category.
A two-stage latent subspace optimization framework was thus proposed, stage one used few-shot samples as positive anchors of the novel class and adjusted the latent space to produce the corresponding results, the second stage governed the generating process via refining the latent space of the unseen categories. 
Through the two-stage optimization, the latent space of the unseen categories can be refined and the generation ability of the latent codes were elevated.
Although inversion-based approaches enable more diverse and controllable image generation, the transformation directions of attributes in the latent space still remain mysterious and require further investigation.

\noindent \textbf{Diffusion-based approaches.}
Inspired by recent remarkable advances of denoising diffusion probabilistic models (DPMs) in visual synthesis, Giannone \etal presented a few-shot diffusion models (FSDM)~\cite{giannone2022few}.
By leveraging conditional DDPMs and vision transformer (ViT)~\cite{ViT, SwinViT}, FSDM can capture image distribution at the image and patch level, enabling the generation of novel images given as few as 5 unseen images at test time.
As the first attempt to integrate DPMs into few-shot image generation, FSDM can inspire more interesting works that utilize the properties of DPMs to improve synthesis quality.

\subsection{Benchmarks and Performances}
\label{sec:benchmarks_few_shot_image_generation}
\noindent \textbf{Datasets.}
There are three popular benchmarks for evaluating the performance of few-shot image generation, namely Flower~\cite{flowers}, Animal Faces~\cite{liu2019few}, and VGGFace~\cite{cao2018vggface2}.
Each of these datasets is divided into seen categories $\mathbb{C}_s$ and unseen categories $\mathbb{C}_u$, and the model is trained on $\mathbb{C}_s$ and tested on $\mathbb{C}_u$.
~\cref{tab:fsig_dataset} presents the detailed splits of these datasets.
The quantitative metrics are computed between the synthesized samples for the unseen categories and the real images from the unseen categories, and the seen categories are only used in the training process. 
These benchmarks provide a standardized way to evaluate the performance of few-shot image generation models.

\begin{table}[h]
\caption{The splits of seen/unseen images (``img'') and classes (``cls'') on three popular datasets for few-shot image generation.} 
\centering
\begin{tabular}{l|lr|lr}  
\toprule[0.8pt]
\multirow{2}{*}{Dataset} & \multicolumn{2}{c|}{Seen} &\multicolumn{2}{c}{Unseen}\cr
                         &\#cls  & \#img    &\#cls  & \#img \cr \hline
            Flowers      & 85    & 3400     & 17    & 680   \cr
            Animal Faces & 119   & 11900    & 30    & 3000  \cr
            VGGFace      & 1802  & 180200   & 552   & 55200 \cr
\bottomrule[0.8pt]
\end{tabular}
\label{tab:fsig_dataset}
\end{table}

\noindent \textbf{Performance.}
~\cref{tab:performance_metric_fsig} presents the FID and LPIPS scores of previous few-shot image generation models on the three widely used datasets.
The ``3-shot" and ``1-shot" settings indicate the number of input conditional images used during model training and testing.
These results demonstrate that:
1) Fusion-based approaches consistently achieve better FID scores compared with other types of methods, suggesting that fusion-based models perform better in synthesis fidelity;
2) Transformation-based methods, which capture the transformation between intra- and inter-categories, can produce more diverse images, identified by higher LPIPS scores that measure the differences between the generated images;
and
3) These models have made significant progress on these datasets, \emph{e.g.,} the FID score of WaveGAN on VGGFace is 4.96. 
However, the resolutions of these datasets (128 $\times$ 128 $\times$ 3) are relatively low, limiting their applications in some practical domains that require high-resolution images.
Accordingly, benchmarks with higher resolution and more objects might further promote the advancement of this field.

\setlength{\tabcolsep}{4pt}
\begin{table*}[t]
\caption{Comparisons of FID ($\downarrow$) and LPIPS ($\uparrow$) scores on images generated by different methods for unseen categories.
FID ($\downarrow$) and LPIPS scores of previous few-shot image generation approaches the three popular benchmarks.
The quantitative metrics are evaluated between generated images and real images of unseen classes. 
The results are quoted from the published papers.
} 
\centering
\resizebox{1.8\columnwidth}{!} {
\begin{tabular}{l|c|c|rr|rr|rr}
\toprule
\multirow{2}{*}{Method} & \multirow{2}{*}{Type} & \multirow{2}{*}{Setting} & \multicolumn{2}{c|}{Flowers} & \multicolumn{2}{c|}{Animal Faces} & \multicolumn{2}{c}{VGGFace}  \cr
& & & FID ($\downarrow$)  & LPIPS ($\uparrow$) &FID ($\downarrow$) & LPIPS ($\uparrow$) & FID ($\downarrow$)  & LPIPS ($\uparrow$)  \cr 
\cmidrule(r){1-1} \cmidrule(r){2-2} \cmidrule(r){3-3} \cmidrule(r){4-5}  \cmidrule(r){6-7}  \cmidrule(r){8-9} 
FIGR~\cite{clouatre2019figr}    & Optimization & 3-shot & 190.12 & 0.0634 & 211.54  & 0.0756 & 139.83 & 0.0834 \cr
GMN~\cite{bartunov2018few}      & Fusion & 3-shot & 200.11 & 0.0743 & 220.45  & 0.0868 & 136.21 & 0.0902 \cr
DAWSON~\cite{liang2020dawson}   & Optimization & 3-shot & 188.96 & 0.0583 & 208.68  & 0.0642 & 137.82 & 0.0769 \cr
DAGAN~\cite{antoniou2017data}   & Transformation & 3-shot & 151.21 & 0.0812 & 155.29  & 0.0892 & 128.34 & 0.0913 \cr
MatchingGAN~\cite{matchinggan}  & Fusion & 3-shot & 143.35 & 0.1627 & 148.52  & 0.1514 & 118.62 & 0.1695 \cr
F2GAN~\cite{hong2020f2gan}      & Fusion & 3-shot & 120.48 & 0.2172 & 117.74  & 0.1831 & 109.16 & 0.2125 \cr
DeltaGAN~\cite{hong2020deltagan}& Transformation & 3-shot & 104.62 & 0.4281 & 87.04   & 0.4642 & 78.35  & 0.3487 \cr
FUNIT~\cite{liu2019few}         & Transformation & 3-shot & 100.92 & 0.4717 & 86.54   & 0.4748 & -      & -      \cr
DiscoFUNIT~\cite{hong2022few}   & Transformation & 3-shot & 84.15  & 0.5143 & 66.05   & 0.5008 & -      & -      \cr
SAGE~\cite{ding2023stable}      & Optimization & 3-shot & 41.35  & 0.4330 & 27.56   & 0.5451 & 32.89  & 0.3314 \cr
LoFGAN~\cite{lofgan}            & Fusion & 3-shot & 79.33 & 0.3862 & 112.81   & 0.4964 & 20.31  & 0.2869 \cr
AMMGAN~\cite{li2023ammgan}                  & Fusion & 3-shot & 75.40  & 0.3990 & 105.11   & 0.5123 & 40.22  & 0.2987 \cr
WaveGAN~\cite{wavegan}          & Fusion & 3-shot & 42.17 & 0.3868 & 30.35   & 0.5076 & 4.96  & 0.3255 \cr
LSO~\cite{zheng2023my}                     & Inversion & 3-shot & 34.59 & 0.3914 & 23.67   & 0.5198 & 3.98  & 0.3344 \cr
\midrule
DAGAN~\cite{antoniou2017data}   & Transformation & 1-shot & 179.59 & 0.0496   & 185.54 & 0.0687  & 134.28  & 0.0608 \cr
DeltaGAN~\cite{hong2020deltagan}& Transformation & 1-shot & 109.78 & 0.3912   & 89.81  & 0.4418  & 80.12  & 0.3146 \cr
FUNIT~\cite{liu2019few}         & Transformation & 1-shot & 105.65 & 0.4221   & 88.07  & 0.4362  & -  & - \cr
Disco-FUNIT~\cite{saito2020coco}& Transformation & 1-shot & 90.12  & 0.4436   & 71.44  & 0.4411  & -  & - \cr
LoFGAN~\cite{lofgan}            & Fusion & 1-shot & 137.47 & 0.3868 & 152.99   & 0.4919 & 26.89  & 0.3208 \cr
AGE~\cite{ding2022attribute}    & Inversion & 1-shot & 45.96 & 0.4305 & 28.04   & 0.5575 & 34.86  & 0.3294 \cr
WaveGAN~\cite{wavegan}          & Fusion & 1-shot & 55.28 & 0.3876 & 53.95   & 0.4948 & 12.28  & 0.3203 \cr
HAE~\cite{li2022euclidean}      & Inversion & 1-shot & 64.26 & 0.4739 & 28.93   & 0.5417 & 35.93  & 0.5919 \cr
LSO~\cite{zheng2023my}          & Inversion & 1-shot & 35.87 & 0.4338 & 27.20   & 0.5382 & 4.15  & 0.3834 \cr
\bottomrule
\end{tabular}
}
\label{tab:performance_metric_fsig}
\end{table*}

\section{One-shot Image Generation}
\label{sec:one_shot_geneartion}
In this section, we first provide the definition of one-shot image generation in~\cref{sec:definition_one_shot_image_geneartion}, then we provide our taxonomy on existing approaches for one-shot image generation in~\cref{sec:taxonomy_one_shot_image_geneartion}, and finally, we present the quantitative metrics and compare the quantitative and qualitative performances of existing models on popular benchmarks in~\cref{sec:benchmark_one_shot_image_geneartion}.

\subsection{Problem Definition}
\label{sec:definition_one_shot_image_geneartion}
One-shot image generation refers to the task of training a generative model to produce novel and diverse images using only a single reference image, without the use of any pre-trained generative models for knowledge transfer. 
This task is of significant importance as it demonstrates the potential application of generative models in practical domains where collecting large-scale training samples is not feasible.
Consequently, the model is expected to capture the internal distribution of the training image and generate diverse images that share the same internal distribution as the reference image, which is an extremely challenging task.
Intuitively, synthesizing images directly from only one image presents a risk of low-variation generation, as the model may simply replicate the given sample.
However, existing methods address this issue by modeling the internal statistics of patches within the training image, allowing the model to capture the information of the target distribution.
~\cref{fig:problem_definition} presents the classical framework of one-shot generative models, commonly employing a multi-scale training architecture.
In the following context, popular solutions for one-shot image generation and their characteristics will be introduced.  
%

\subsection{Model Taxonomy}
\label{sec:taxonomy_one_shot_image_geneartion}
According to whether training is required for generating novel images, existing one-shot generative models can be broadly two categories: non-parametric approaches and parametric approaches.
Non-parametric approaches require model training to capture the internal distribution of each image, while parametric approaches employ classical patch-based nearest neighbor matching to produce new images.
Model training based approaches could be further categorized into GAN-based and diffusion-based according to which generative model is used for training.
The details of these methods are presented and discussed below.

\noindent \textbf{GAN-based approaches.}
The idea of learning a generative model from a single image was first proposed in SinGAN~\cite{singan}, which introduced a multi-scale GAN framework that captures the internal distribution of patches within the input image in a coarse-to-fine manner.
At each scale $n$, the model generated an image by upsampling the previous scale's image through $G_n$, and a random noise was added to the input of $G_n$.
Moreover, the discriminator at the $n$-th scale was responsible for distinguishing the patches in the down-sampled training image, $x_n$, and the generated image $G_n(z_n, \hat{x}_{n-1})$.
such a multi-scale pipeline formulated a pyramid of GANs, and these subnetworks were trained sequentially from coarsest to finest scale.
Once trained, the model can hierarchically generate novel images.
Recurrent SinGAN~\cite{he2021recurrent} replaced the pyramid of generators in SinGAN with a single recurrent generator, enabling a scale-agnostic one-shot generative model.
In order to improve the generator's ability to capture the global structure of the training image, Chen \etal imported a self-attention mechanism to SinGAN~\cite{chen2021sa}.
However, training a pyramid of GANs sequentially was time-consuming, and ConSinGAN~\cite{hinz2021improved} proposed some best practices to improve the synthesis quality beyond that of SinGAN.
Concretely, ConSinGAN concurrently trained several stages in a sequential multi-stage way, significantly improving the training speed and enabling fewer parameters.
Furthermore, ExSinGAN~\cite{zhang2021exsingan} combined external prior obtained by GAN inversion with the information of internal patches to obtain better generation quality and competitive generalization ability for manipulating the input image.

Unlike the two-stage approaches that first train the generator on low-resolution images and then optimize for high-resolution generation, Zhang \etal proposed PetsGAN~\cite{zhang2022petsgan}, which leveraged the external and internal priors in one stage.
In particular, the external priors were obtained via a lightweight deep external prior network, providing high-level information for generation, while the internal priors reduced the patch discrepancy between the synthesized image and the training image using a global reconstruction loss.
PetsGAN was trained in an end-to-end manner with strong priors, thus effectively speeding up the training process and achieving performance improvements.
Similarly, Sushko \etal developed an end-to-end one-shot GAN that could learn to generate samples from one image or one video~\cite{sushko2021one}.
The key component of their one-shot GAN was a two-branch discriminator with content and layout branches that respectively discriminate the internal content and the scene layout realism.
Such designs allowed a realistic and diverse generation with varying content and layout.
Recently, Jiang \etal pointed out that existing CNN-based one-shot GANs struggled to extract and maintain the global structural information~\cite{jiang2023tcgan}.
Accordingly, they exploited vision transformer (ViT)~\cite{ViT, SwinViT} to capture the global structure of an image and maintain the integrity of semantic-aware information.
Together with a scaling formula that had scale-invariance, their proposed model TcGAN effectively improved the quality of image generation and super-resolution tasks.

Although existing GAN-based one-shot generative models have demonstrated impressive abilities in diverse generation and manipulating input images, they require re-training for any new input images, which is time-consuming and expensive.
Approaches that can 1) quickly capture the internal distribution and 2) generalize to new samples are more favorable for practical applications.

\noindent \textbf{Diffusion-based approaches.}
Denoising diffusion probabilistic models (DPMs) have become the most popular generative model in the community, achieving unprecedented improvements in image synthesis. 
As an emerging research topic in the last two years, there have been efforts to investigate the performance of DPMs in capturing the internal distribution of a single image.
Wang \etal designed a patch-wise denoising framework dubbed SinDiffusion ~\cite{wang2022sindiffusion} with two key ingredients.
First, in order to avoid the accumulation of errors and reduce the training cost, SinDiffusion was trained with a single model to progressively generate images over timesteps.
Second, SinDiffusion learned to estimate the noise based on a local patch instead of the whole training image, enabling the model to have a reasonable receptive field for diverse output.
Once trained, SinDiffusion could be applied to a variety of image manipulation tasks, including text-guided image generation, image outpainting, image harmonization, and more. 
SinDDM~\cite{kulikov2023sinddm} was a concurrent work that learned the internal statistics of the input image with a multi-scale diffusion process.
A convolutional lightweight denoiser conditioned on the noise level and the scale was proposed to derive the reverse diffusion process for training.
In this way, SinDDPM was capable of producing novel samples in a coarse-to-fine way with arbitrary dimensions conditioned on various scales and different timesteps.
Similarly, Nikankin \etal proposed SinFusion~\cite{nikankin2022sinfusion}, which modeled the appearance and dynamics of a single input image or video.
Particularly, SinFusoin trained on large crops (~95\%) from a single image to preserve the overall structure and appearance of the input image.
Moreover, ConvNext~\cite{liu2022convnet} blocks were employed to replace the functionality of the attention layers in the diffusion UNet, reducing the receptive field.
SinFusion can be applied uniformly to various tasks of single image and video editing, which were not accomplished in previous works.

Despite the significant advancements in image quality and manipulation ability achieved by diffusion-based one-shot generation models, some limitations remain underexplored. 
Specifically, the iterative reverse process employed in DPMs leads to slow training and inference times, which can limit the practicality of these models for real-time applications. 
Additionally, the internal statistics of the training image in DPMs are often less confined, which can lead to instability in the training process and hinder the model's ability to generalize to new images.
To address these limitations, future research efforts could be directed toward designing faster inference techniques for DPMs and incorporating suitable priors to improve the stability of the training process and the generalization ability of the model.

\noindent \textbf{Non-parametric approaches.}
Despite the impressive performance of parametric-based generative models, they often require long training times and can suffer from unsatisfactory artifacts.
In contrast, patch-based approaches require no costly training and can yield better visual quality with fewer artifacts.
For instance, Granot \etal proposed a generative patch-base algorithm named GPNN~\cite{granot2022drop} that employed non-parametric patch nearest neighbors to replace similar patches with their nearest counterparts and combine multiple patches into a novel image.
GPNN was free of training and ensured that each pixel of the new image was adopted from the training image, resulting in improved visual quality.
In order to further enhance the fidelity of patch-based models, Cherel \etal developed an initialization scheme based on optimal transport and minimization of a patch energy~\cite{cherel2022patch} that respected the patch distribution of the training image and encourages diversity.
Notably, they found that choosing proper initialization for patch-based models was crucial for diversity.
Similarly, Elnekave \etal used the sliced Wasserstein distance (SWD)~\cite{pitie2005n, bonneel2015sliced} to compare the distributions of patches through an unbiased estimate of the SWD~\cite{elnekave2022generating}.
Such estimation helped to explicitly and efficiently minimize the distance between internal patch distributions of two images (\emph{i.e.,} the training image and the generated one), enabling more plausible generation in just a few seconds.
Albeit computationally simple and effective, non-parametric models suffer from low diversity because the synthesized images can be viewed as rearrangements of the internal patches of the referenced image.
Additionally, Additionally, when applied to images with a global coherent structure, such as a human face or a chair, non-parametric models may produce corrupted global structures and artifacts.
To mitigate these issues, integrating additional structural information into the patch-based matching process could be a promising solution.

\subsection{Benchmarks and Performances}
\label{sec:benchmark_one_shot_image_geneartion}
\begin{figure}
    \centering
    \includegraphics[width=\linewidth]{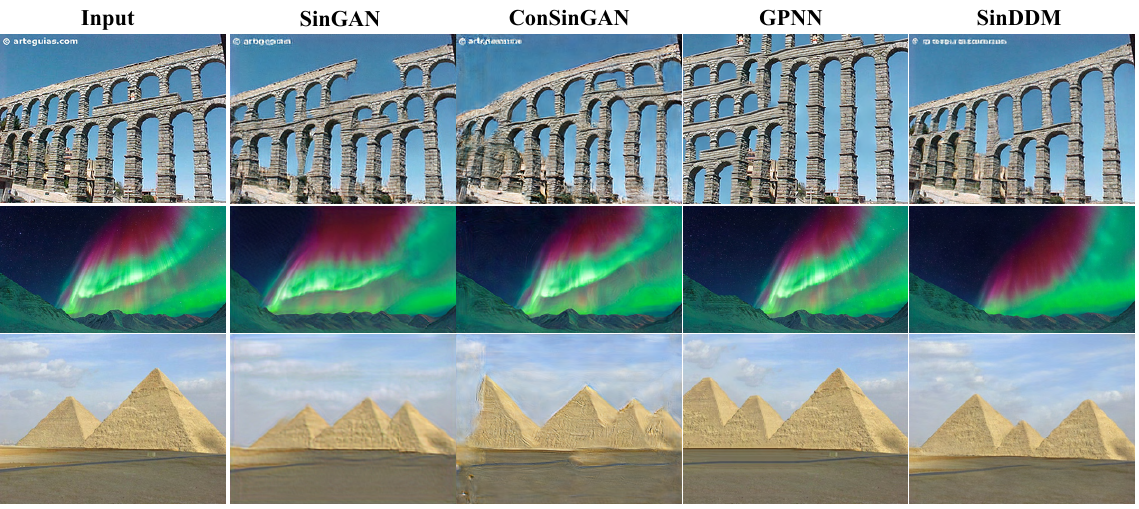}
    \caption{
    Qualitative comparisons of different one-shot generative models.
    Images quoted from~\cite{kulikov2023sinddm}.
    }
    \label{fig:comparison_one_shot}
\end{figure}

\noindent \textbf{Quantitative metrics.}
The FID~\cite{fid} is a common metric used to evaluate the performance of generative models by quantifying the distributional discrepancy between the synthesized and observed distribution.
However, in a one-shot setting where only on single training image is available, it is more important to measure the internal distributional divergences of the generated images.
To address this, Single Image FID (SIFID)~\cite{singan} was proposed to compute the internal distribution distances, providing a more appropriate evaluation metric for one-shot generation tasks.
In addition to SIFID, pixel-wise and sample-wise diversity can also be computed to investigate the synthesis diversity of the generated images. 
Furthermore, several no-reference image quality measures, including NIQE~\cite{mittal2012making}, NIMA~\cite{talebi2018nima}, and MUSIQ~\cite{ke2021musiq}, can also be employed to evaluate the quality of generated images.
These metrics provide complementary information to SIFID and can provide a more comprehensive evaluation of the quality and diversity of the generated images.

\noindent \textbf{Performances.}
Notably, since only a single input image is required for one-shot image generation, each image could be used as the benchmark for qualitative and quantitative evaluation.
~\cref{tab:quantitative_one_shot} provides the quantitative results for one-shot image generation.
Particularly, each measure in the table is computed across 50 synthesized images for every training image, and their training and testing protocols are consistent, which is crucial for a fair comparison.
Moreover, ~\cref{fig:comparison_one_shot} shows the qualitative comparison between popular one-shot image generation models.
The quantitative and qualitative results consistently demonstrate that 
1) Approaches such as SinDDM~\cite{kulikov2023sinddm} and ConSinGAN~\cite{hinz2021improved} that require model training perform better in terms of image quality and diversity, as indicated by higher diversity and IQA metrics;
and 
2) Patch-based models achieve better patch-level distribution matching, achieving higher fidelity.
Additionally, these results also suggest that different models have their own strengths and weaknesses, and future research may explore the combination of different approaches to achieve better performance.

\begin{table*}[h]
\centering
\small
\setlength{\tabcolsep}{2pt}
\caption{Quantitative evaluation metrics of different unconditional one-shot image generation models.
These metrics were computed over 50 generated samples per training image and averaged by 12 images used in~\cite{kulikov2023sinddm}.
The results are quoted from the publisheh papers.
} 
  \begin{tabular}{c c c c c c} 
    \toprule
    Type & Metric & SinGAN~\cite{singan} & ConSinGAN~\cite{hinz2021improved} & GPNN~\cite{elnekave2022generating} & SinDDM~\cite{kulikov2023sinddm}\\
    \midrule
    \multirow{2}{*}{Diversity} & Pixel Div. $\uparrow$ & {0.28$\pm$0.15} & 0.25$\pm$0.20 & 0.25$\pm$0.20 & \textbf{{0.32$\pm$0.13}}\\ 
     & LPIPS Div. $\uparrow$ & {0.18$\pm$0.07} & 0.15$\pm$0.07 & 0.10$\pm$0.07 & {{0.21$\pm$0.08}} \\
    \hline
    \multirow{3}{*}{No reference IQA} & NIQE $\downarrow$ & 7.30$\pm$1.50 & {{6.40$\pm$0.90}} & 7.70$\pm$2.20 & {7.10$\pm$1.90}\\ 
     & NIMA $\uparrow$ & {5.60$\pm$0.50} & 5.50$\pm$0.60 & {5.60$\pm$0.70} & {{5.80$\pm$0.60}} \\
     & MUSIQ $\uparrow$ & 43.00$\pm$9.10 & 45.60$\pm$9.00 & {{52.80$\pm$10.90}} & {48.00$\pm$9.80}  \\ 
    \hline
    Patch Distribution & SIFID $\downarrow$ & 0.15$\pm$0.05 & {0.09$\pm$0.05} & {{0.05$\pm$0.04}} & 0.34$\pm$0.30\\ 
    \bottomrule
  \end{tabular}
  \label{tab:quantitative_one_shot}
\end{table*}

\section{Applications and Future Directions}
\label{sec:applications_future_direction}
\begin{figure*}
    \centering
    \includegraphics[width=\linewidth]{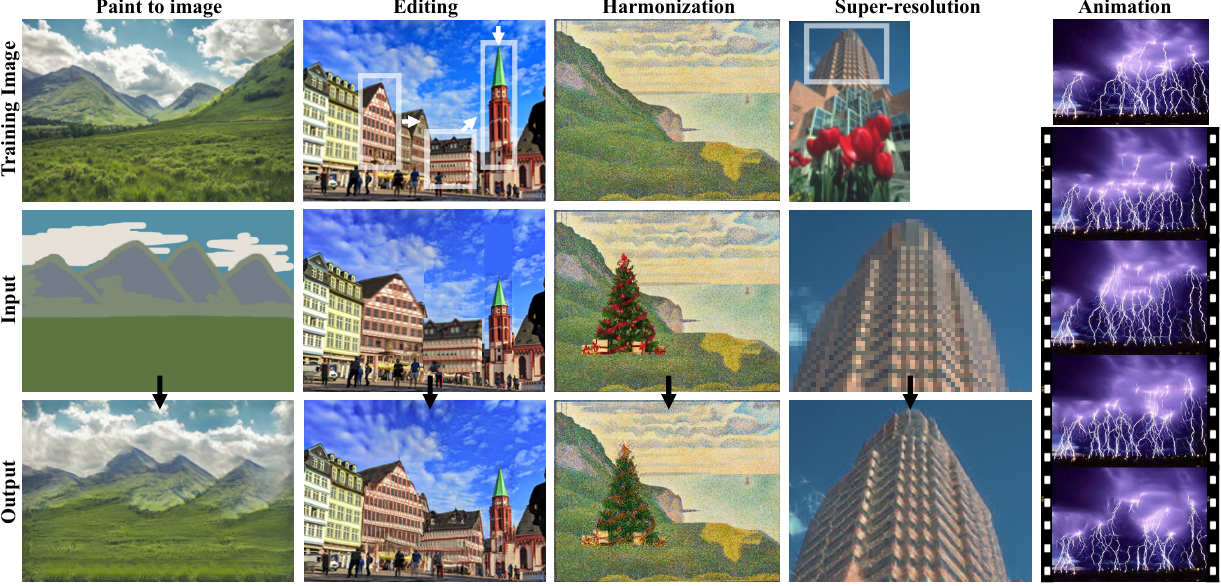}
    \caption{Generative models under limited data have enabled downstream applications in various image editing tasks, such as paint-to-image, image harmonization, super-resolution, and animation.
    Images quoted from~\cite{singan}.
    }
    \label{fig:applications_manipulation}
\end{figure*}

The development of generative models under limited data has spawned many promising applications, as shown in~\cref{fig:applications_manipulation}.
In this section, we first discuss the related applications of generative models under limited data, including image manipulation, stylization, and augmentation for downstream tasks.
However, there is still ample room for further improvement, and we thus highlight several future directions in term of content controllability and editability, evaluation metrics, and other practical applications.

\subsection{Applications}

\noindent \textbf{Image editing/Manipulation.}
In addition to generating new images from randomly sampled noises, existing generative models under limited data have enabled various applications for image editing and manipulation, as illustrated in~\cref{fig:applications_manipulation}.
Ideally, once trained, the learned model could be leveraged for manipulating the generated images with additional control conditions as input, such as low-resolution images and editing masks, by borrowing the generation ability.
For instance, SinFusion~\cite{nikankin2022sinfusion} can generate novel images given only a single image and sketch conditions as input, demonstrating its potential in image manipulation tasks.
Moreover, novel images could be obtained by interpolating/mixing the latent codes~\cite{shen2020interfacegan, shen2020interpreting} without user-defined input, providing an additional level of control over the synthesized images.

\begin{figure}
    \centering
    \includegraphics[width=\linewidth]{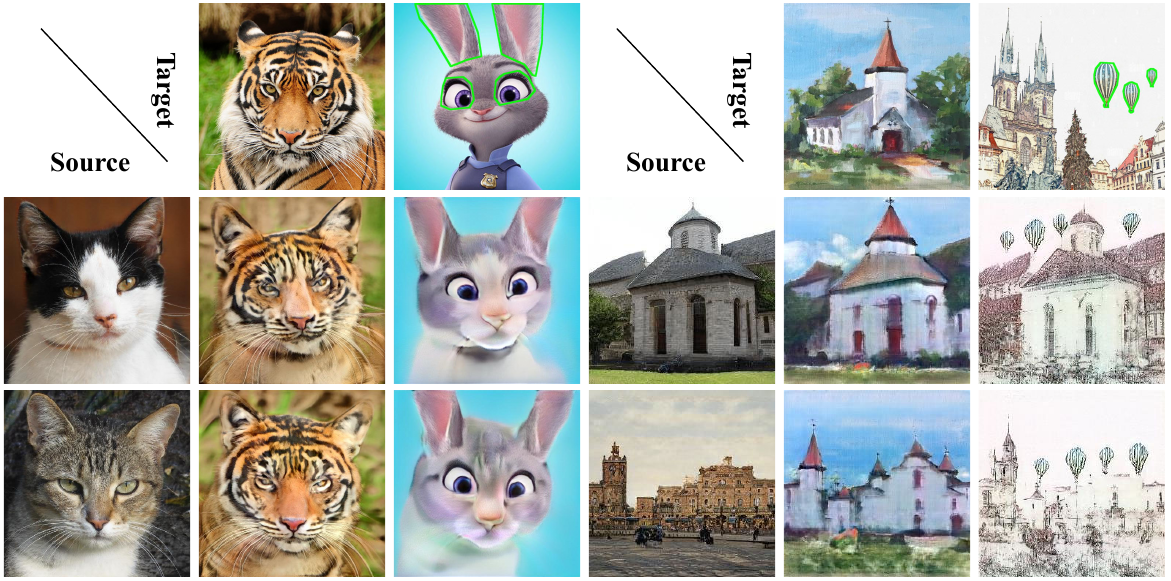}
    \caption{Generative models under limited data can be employed to transfer various source domains to various target domains with different styles.
    Images quoted from~\cite{NEURIPS2022_58ce6a4b}.
    }
    \label{fig:applications_artilization}
\end{figure}

\noindent \textbf{Artilization and Stylization.}
Generative models under limited data also enjoy the potential to assist artists in their creative process by producing novel concepts and inspiring new ideas.
Specifically, an artist can generate various samples using a generative model trained on only several images of their artworks, providing them with fresh perspectives and ideas for future creations.
Additionally, users can stylize their photos with these generative models, as shown in~\cref{fig:applications_artilization}, transforming their photos into unique and artistic pieces.
This has significant potential in various domains, such as social media and the creative industry, where visually appealing content is highly valued.

\begin{figure}
    \centering
    \includegraphics[width=\linewidth]{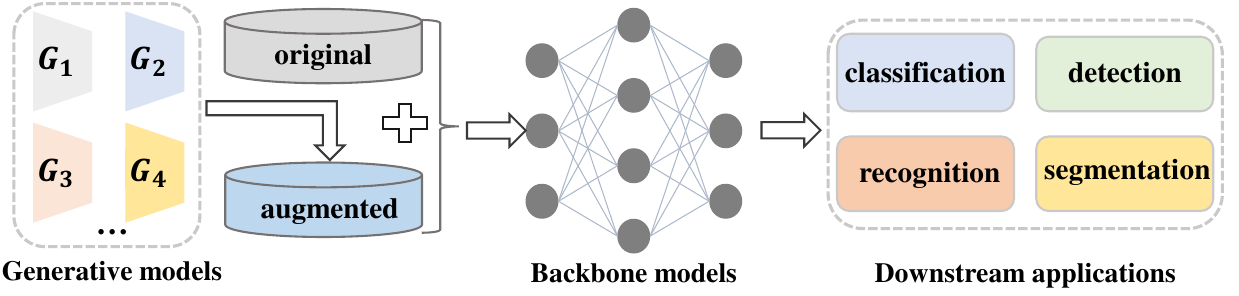}
    \caption{Generative models under limited data could be employed to produce novel samples to augment the original training sets, facilitating the performance of various downstream tasks, including image classification, object detection/recognition, and semantic segmentation. 
    }
    \label{fig:applications_augmentation}
\end{figure}

\noindent \textbf{Augmentation for downstream tasks.}
Considering the trained generative models can produce novel images, it is intuitive to employ them to generate new samples for downstream tasks, like classification, detection, and segmentation tasks.
The overall pipeline of enlarging the training sets with generative models is given~\cref{fig:applications_augmentation}.
Several works have demonstrated that using synthesized images can indeed facilitate downstream applications~\cite{matchinggan, hong2020f2gan, wavegan, ding2022attribute}, providing an alternative for various tasks, particularly when the training data distribution is small-scale or imbalanced.

\subsection{Future Directions}

\noindent \textbf{Controllability and editability.}
Despite the fascinating features of image editing and manipulation enabled by prior works, there is still a significant gap with practical applications regarding the controllability and editability of the generated images.
In particular, users prefer a friendly interactive interface for controlling the generated content and editing the local details of the generated images.   
Additionally, it would be desirable to allow controlling generated content via multi-modal signals, such as text- and speech-conditioned instructions.
Improving the interactivity of the synthesis models is an appealing direction for addressing these challenges. 
By enhancing the user interface and incorporating multi-modal signals, generative models can provide a more intuitive and flexible way for users to interact with the generated images and achieve their desired outcomes.

\noindent \textbf{New metrics.}
Existing evaluation metrics have been identified with several flaws, highlighting the need for new metrics that can more precisely and reliably reflect the generation quality.
For instance, studies~\cite{kynkaanniemi2022role, Trend, yang2023revisiting} have pointed out that due to the existence of a large perceptual null space, the FID metric could be altered without any change to the generator, leading to unreliable evaluations.
Besides, the sample efficiency of FID is relatively poor, as FID scores fluctuate greatly with respect to different amounts of evaluated samples~\cite{yang2023revisiting}.
Although KID~\cite{KID} has been identified to be more reliable under limited data settings~\cite{stylegan2-ada}, its correlation with human visual judgment is still under-explored. 
Therefore, developing new metrics that could
1) precisely and reliably reflect the generation quality,
2) consistently measure the synthesis under various amounts of samples, 
and 3) agree with human visual perception
is crucial for the community.
Moreover, future research may explore the use of subjective evaluation metrics, such as human perceptual studies, to complement traditional objective metrics.
This can provide a more holistic evaluation of the generative models' performance and enable a better understanding of the model's strengths and weaknesses.

\noindent \textbf{Correlation between generative modeling and representation learning.}
Generative modeling and representation learning are two vital branches in the field of computer vision, and they are often studied independently in their respective paradigms.
However, they share similar high requirements in terms of representation at the instance level, with generative models needing to produce novel images and representation learning models needing to extract representative features for downstream applications.
Therefore, it is a promising direction to train image generation (\emph{e.g.,} GANs, Diffusions), and representation learning models in a unified framework is a promising direction that could enable cooperation between these two branches.
In this way, generative models and discriminative models to better contribute to each other, which is especially meaningful in the few-shot regimes where the amount of training data is limited.

\noindent \textbf{Personalized services.}
Producing novel images of one's own images, such as pet cats, with various shapes/gestures in different scenarios given few input images is a challenging yet practically useful editing task. 
This task requires the model to preserve the identity while producing various fine details that are harmonious in each pixel of the synthesized images.
Recent approaches, such as~\cite{hertz2022prompt, ruiz2023dreambooth, yang2023controllable}, address this task mostly via leveraging the appealing synthesis ability of large-scale pre-trained text-to-image diffusion models, \emph{e.g.,} Stable Diffusion.
However, the synthesis quality and inference speed of these models can still be improved.
Additionally, it is worth investigating whether there are any opportunities to train from scratch on the few input images to provide customized services.

\noindent \textbf{Training stability.}
Despite tremendous efforts have been poured to ameliorate the training dynamics of generative models under limited data, the issues of model overfitting and memorization remain prevalent, particularly under one-shot settings, where only a single training example is available.
Diffusion models are emerging as a new trend in the field of generative models due to enhanced stability during training and relatively unexplored performance in the limited generation scenarios,
However, their performances in the field of limited generation are relatively less explored. 
Therefore, incorporating diffusion models and investigating beneficial attributes of them to help stabilize the training process are practical directions for future research.

\section{Conclusion}
\label{sec:conclusion}
In this survey, we present a comprehensive overview of image synthesis models under limited data and categorize existing research in this area into four sub-tasks: data-efficient generative models, few-shot generative adaptation, few-shot image generation, and one-shot image synthesis.
We summarize various solutions, benchmarks, and performances for these tasks and thoroughly analyze the advantages, disadvantages, and limitations of existing approaches. 
Furthermore, we discuss the potential applications of image synthesis under limited data and identify promising directions for future research.

Image synthesis under limited data enjoys great opportunities in various practical domains, yet faces many challenges simultaneously.
Our survey aims to provide readers with a better understanding of the field of data-efficient synthesis and inspire further research in this area.
Hopefully, our review could provide valuable insights to the community and stimulate further exciting works in the future.

\ifCLASSOPTIONcompsoc
  \section*{Acknowledgments}
\else
  \section*{Acknowledgment}
\fi
This work is supported by Shanghai Science and Technology Program "Distributed and generative few-shot algorithm and theory research" under Grant No. 20511100600 and ``Federated based cross-domain and cross-task incremental learning" under Grant No. 21511100800, Natural Science Foundation of China under Grant No. 62076094, Chinese Defense Program of Science and Technology under Grant No.2021-JCJQ-JJ-0041, China Aerospace Science and Technology Corporation Industry-University-Research Cooperation Foundation of the Eighth Research Institute under Grant No.SAST2021-007.

\ifCLASSOPTIONcaptionsoff
  \newpage
\fi
\bibliographystyle{IEEEtran}
\bibliography{references}


\end{document}